%% file: main_ssin.tex
\documentclass[acmsmall,usenames,dvipsnames,table,xcdraw]{acmart}

\usepackage{enumitem}
\usepackage{threeparttable}
\usepackage[caption=false, font=footnotesize]{subfig}
\usepackage{hyperref}

\usepackage{multirow}
\usepackage[normalem]{ulem}
\useunder{\uline}{\ul}{}

\usepackage{stfloats}
\usepackage{booktabs}
\usepackage{amsmath}

\setcopyright{acmlicensed}
\acmJournal{PACMMOD}
\acmYear{2023} \acmVolume{1} \acmNumber{2} \acmArticle{176} \acmMonth{6} \acmPrice{15.00}\acmDOI{10.1145/3589321}

\begin{document}

\title{SSIN: Self-Supervised Learning for Rainfall Spatial Interpolation}

\author{Jia Li}
\email{jlidw@cse.ust.hk}
\orcid{0000-0002-2051-635X}
\affiliation{
  \institution{Department of Computer Science and Engineering, The Hong Kong University of Science and Technology}
  \city{Hong Kong SAR}
  \country{China}
}

\author{Yanyan Shen}
\authornote{The corresponding author.}
\email{shenyy@sjtu.edu.cn}
\orcid{0000-0001-8364-3674}
\affiliation{
  \institution{Department of Computer Science and Engineering, Shanghai Jiao Tong University}
  \city{Shanghai}
  \country{China}
}

\author{Lei Chen}
\email{leichen@cse.ust.hk}
\orcid{0000-0002-8257-5806}
\affiliation{
  \institution{Department of Computer Science and Engineering, The Hong Kong University of Science and Technology}
  \city{Hong Kong SAR}
  \country{China}
}

\author{Charles Wang Wai Ng}
\email{cecwwng@ust.hk}
\orcid{0000-0001-6693-3151}
\affiliation{
  \institution{Department of Civil and Environmental Engineering, The Hong Kong University of Science and Technology}
  \city{Hong Kong SAR}
  \country{China}
}

\renewcommand{\shortauthors}{Jia Li, Yanyan Shen, Lei Chen, \& Charles Wang Wai Ng}

\input{0-Abstract}

\ccsdesc[500]{Mathematics of computing~Interpolation}
\ccsdesc[500]{Information systems~Geographic information systems}

\keywords{spatial interpolation; self-supervised learning; transformer}

\maketitle

\input{1-Introduction}
\input{2-Related_work}

\input{3-Methodology}
\input{4-Experiments}
\input{5-Conclusions}

\begin{acks}
Yanyan Shen's work is supported by 
the National Key Research and Development Program of China (2022YFE0200500), 
Shanghai Municipal Science and Technology Major Project (2021SHZDZX0102) 
and SJTU Global Strategic Partnership Fund (2021 SJTU-HKUST).
Lei Chen’s work is supported by 
National Science Foundation of China (NSFC) under Grant No. U22B2060,
the Hong Kong RGC GRF Project 16209519, 
CRF Project C6030-18G, C2004-21GF, 
AOE Project AoE/E-603/18, 
RIF Project R6020-19, 
Theme-based project TRS T41-603/20R, 
China NSFC No. 61729201, 
Guangdong Basic and Applied Basic Research Foundation 2019B151530001, 
Hong Kong ITC ITF grants MHX/078/21 and PRP/004/22FX, 
Microsoft Research Asia Collaborative Research Grant, 
HKUST-Webank joint research lab grant and 
HKUST Global Strategic Partnership Fund (2021 SJTU-HKUST).
\end{acks}

\bibliographystyle{ACM-Reference-Format}
\bibliography{references}

\received{October 2022}
\received[revised]{January 2023}
\received[accepted]{February 2023}

\end{document}

%% file: 0-Abstract.tex
\begin{abstract}
The acquisition of accurate rainfall distribution in space is an important task in hydrological analysis and natural disaster pre-warning. 
However, it is impossible to install rain gauges on every corner. 
Spatial interpolation is a common way to infer rainfall distribution based on available raingauge data.
However, the existing works rely on some unrealistic pre-settings to capture spatial correlations, which limits their performance in real scenarios. 
To tackle this issue, we propose the SSIN, which is a novel data-driven self-supervised learning framework for rainfall spatial interpolation by mining latent spatial patterns from historical observation data.
Inspired by the Cloze task and BERT, we fully consider the characteristics of spatial interpolation and design the SpaFormer model based on the Transformer architecture as the core of SSIN.
Our main idea is: by constructing rich self-supervision signals via random masking, SpaFormer can learn informative embeddings for raw data and then adaptively model spatial correlations based on rainfall spatial context.
Extensive experiments on two real-world raingauge datasets show that our method outperforms the state-of-the-art solutions.
In addition, we take traffic spatial interpolation as another use case to further explore the performance of our method, and SpaFormer achieves the best performance on one large real-world traffic dataset, which further confirms the effectiveness and generality of our method.
\end{abstract}

%% file: 1-Introduction.tex
\section{Introduction} \label{sec:Introduction}
Acquiring fine-grained rainfall data in space and time is critical for hydrological studies and early warning of natural disasters.
In many areas, automatic stations have been established to deliver rainfall accumulation data with high temporal resolution (e.g., hourly).
Considering that only station data provide direct rainfall measurements, researchers rely on spatial interpolation techniques to infer fine-grained rainfall in space from sparse station observations.

The goal of performing spatial interpolation is to 
\textit{``predict'' data for any locations with no historical observations based on available station observations.}
Rainfall spatial interpolation faces the following practical challenges:
(1) \textit{Complex Spatial Pattern.} Rainfall usually shows irregular and non-uniform distribution in space. 
(2) \textit{Dynamic-Changing Spatial Patterns.} Rainfall is a dynamic process involving complex spatiotemporal evolution~\cite{hy1}, spatial correlations in rainfall between locations can vary significantly over time. 
(3) \textit{Lack of useful auxiliary variables.} Due to cost constraints, it is non-trivial to obtain enough other observed variables to characterize rainfall spatial patterns for interpolation.
These challenges together make it difficult to implement accurate spatial interpolation for rainfall.

\begin{figure}[t]
\centering
\includegraphics[width=0.95\columnwidth]{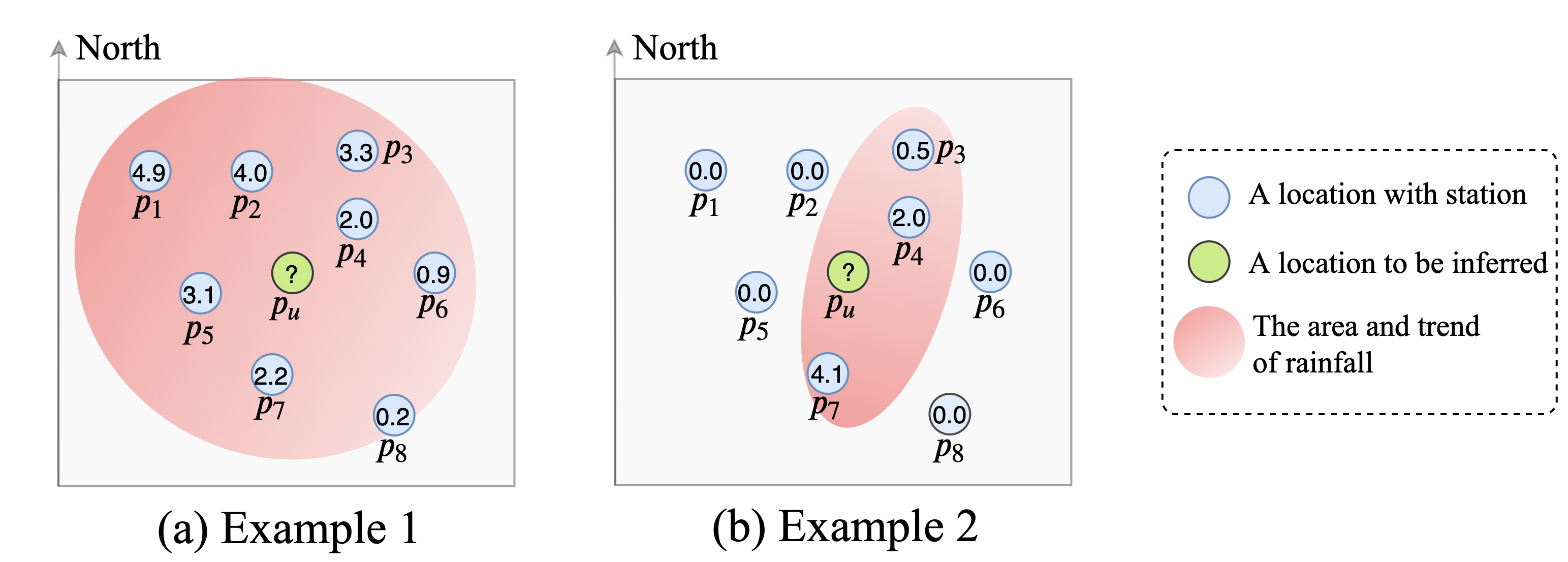}
\caption{The examples of different rainfall distributions.}
\label{img:rain_case}
\end{figure}

Various interpolation methods have been applied to the rainfall spatial interpolation task.
Traditional methods~\cite{hy2,hy3,hy4,hy5,hy6} formulate spatial interpolation as a linear weighted sum of observed values to estimate the values at unobserved locations,
which can be generally classified into two categories: (i) deterministic approaches, such as IDW~\cite{hy5} and TIN~\cite{hy4}; (ii) geostatistical approaches, such as Kriging~\cite{hy6}.
Recently, with the development of Graph Neural Networks (GNNs)~\cite{ml32}, researchers have proposed GNN-based models to handle spatial interpolation tasks by modeling spatial points as a graph, such as KCN~\cite{ml5} and IGNNK~\cite{ml16}.
Although existing techniques are effective to some extent, they still suffer from an intrinsic restriction --- relying on various pre-settings to capture spatial correlations, which will bring two issues:
\begin{itemize}[leftmargin=*]
\item
First, parameter selections in these pre-settings are highly dependent on the researchers' modeling experience, and inappropriate parameters may lead to incorrect estimation of spatial correlation, thereby affecting interpolation performance.
For example, in Kriging~\cite{hy6}, 
the variogram essentially represents the strength of spatial correlations between data points and an inappropriate variogram model can lead to completely false results;
KCN~\cite{ml5} and IGNNK~\cite{ml16} construct the adjacency matrix using a Gaussian kernel based on distance, in which the kernel length is an important parameter that needs tuning for better performance.

\item
Second, even with the best possible parameters, the spatial correlations explicitly characterized by pre-settings may not be well suited for real-world scenarios.
For example, Kriging assumes observation data are from an underlying Gaussian process, which may not hold in real rainfall data.
Besides, most methods tend to use distance-based functions to measure spatial correlations: e.g., IDW uses the function of inverse distance to directly measure the correlation between locations; KCN and IGNNK construct their adjacency matrix using a Gaussian function based on distance to denote spatial correlation and guide message passing.
However, static geographic distances can hardly reflect complex spatial patterns of various rainfall events.
For example, as shown in Figure~\ref{img:rain_case}, given eight stations $p_1-p_8$ and one location $p_u$ to be inferred: while the distances between locations are fixed, the correlations between them may change in different rainfall events --- in Example 1, all nine locations are in the same rainfall field and there is a correlation between any one station and $p_u$; but in a local convective rain event like Example 2, only $p_3$, $p_4$ and $p_7$ have correlations with $p_u$.\footnote{The reality may be more complicated, and the strength of correlations may vary.}
\end{itemize}
Hence, the performance of existing methods is still limited by their inflexible and unrealistic pre-settings.

In this study, we aim to develop an effective spatial interpolation solution to overcome the above issues. 
To address the limitations of pre-settings, one common idea is to adaptively capture the intrinsic correlations from the data itself, which naturally leads us to self-supervised learning.
In recent years, self-supervised learning with Transformer~\cite{ml1} has achieved great success in natural language processing (NLP).
Especially BERT~\cite{ml9} and its proposed Masked Language Model (MLM) are advancing the development of other domains~\cite{ml28,ml29,ml30,ml31}.
MLM is inspired by the Cloze task and its key idea is: randomly mask a proportion of tokens in the input sequence, then train the model to predict the masked tokens based on their context.
We observed that rainfall spatial interpolation is essentially a fill-in-the-blank problem in the spatial domain, which is an ``extended version'' of the Cloze task.
Inspired by the success of BERT and MLM, we propose to borrow the idea of self-supervised learning to solve rainfall spatial interpolation: based on large amounts of historical data, we can employ a similar mask-and-recover task to enable the model to capture the latent spatial correlations from rainfall spatial context. 

\begin{figure}[t]
\centering
\subfloat[The Cloze Task]{
\centering
\includegraphics[width=0.4\columnwidth]{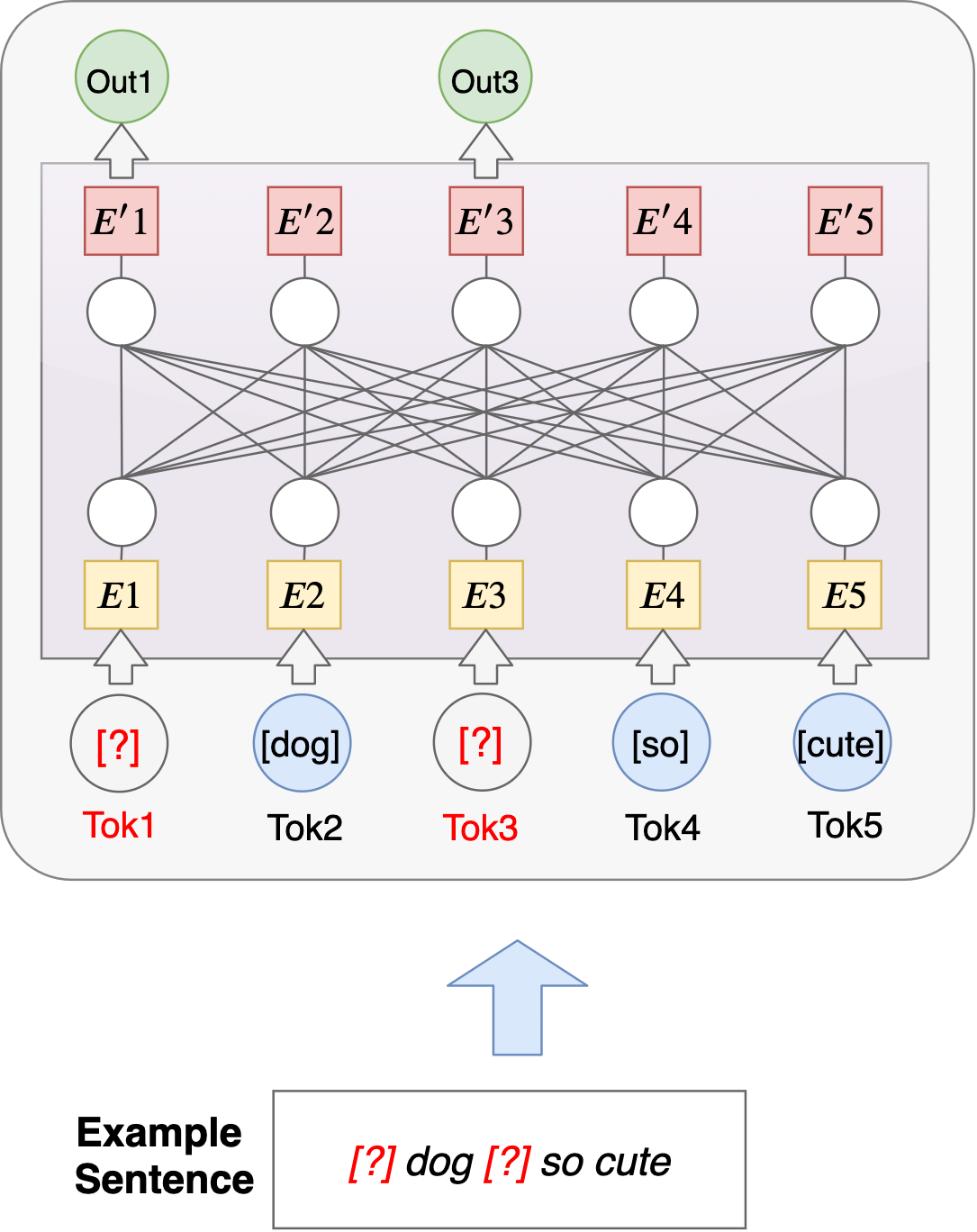}
\label{img:differences_1}
}\hspace{5mm}
\subfloat[Spatial Interpolation]{
\centering
\includegraphics[width=0.412\columnwidth]{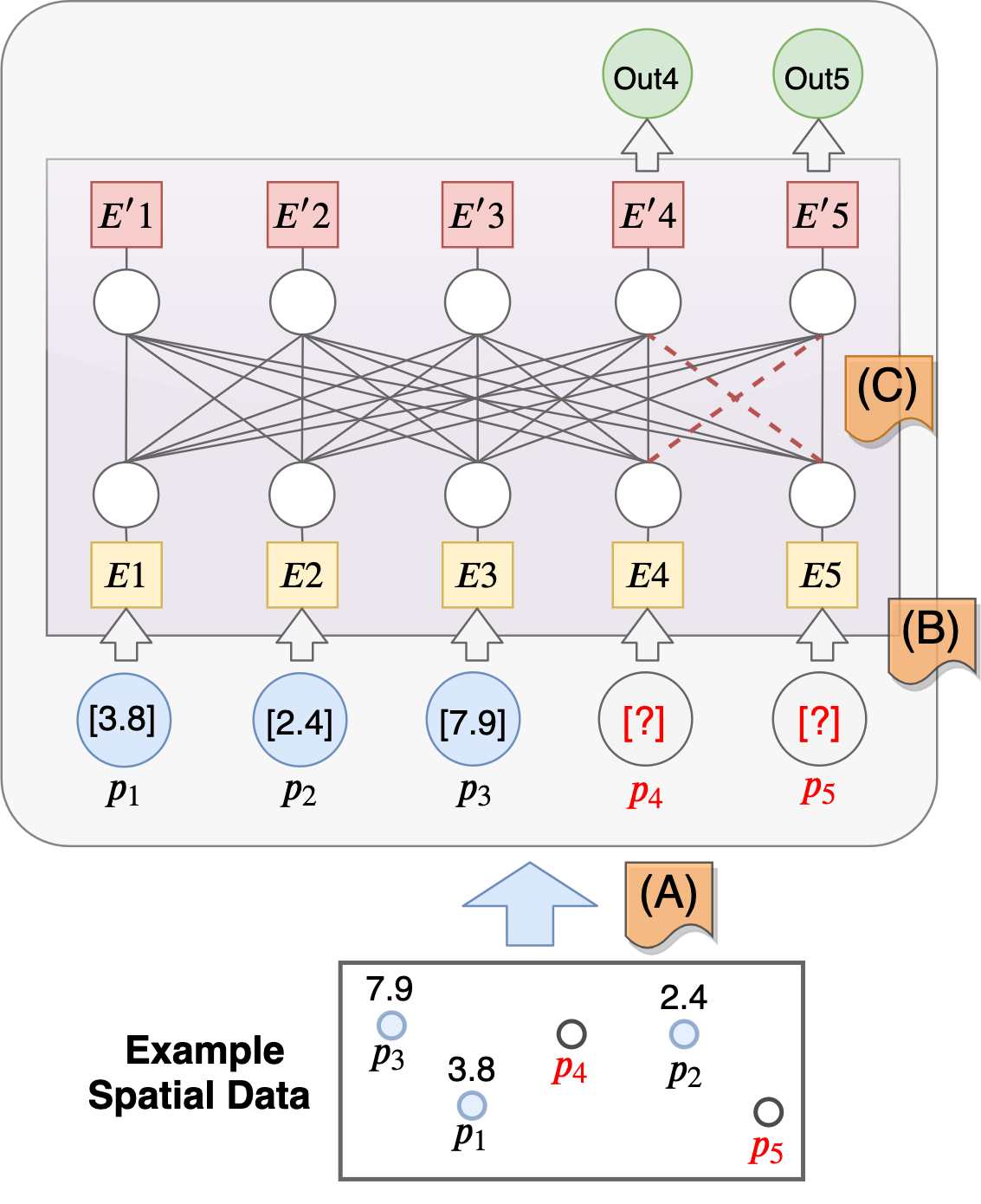}
\label{img:differences_2}
}
\caption{Distinctions between tasks affecting model design.}
\label{img:task_difference}
\end{figure}

However, applying the BERT model directly to spatial interpolation is inappropriate since it is designed for language modeling.
To develop a promising solution, differences between tasks should be well considered. 
As shown in Figure~\ref{img:task_difference}, the key distinctions affecting model design are summarized as follows:
\begin{itemize}[]
\item[(A)]
Unlike language sentences, there is no concept of ``sequence'' in the continuous 2D space. Hence, the model should be extended to handle spatial data.
\item[(B)]
Different from the discrete language data, the observations and positions in spatial interpolation are both continuous. A look-up table manner is no longer suitable for data embeddings since there may be infinite values. Therefore, the model needs to employ new embedding methods for continuous inputs.
\item[(C)]
In the Cloze task, given the known token information, the positions of missing tokens are determined. But in spatial interpolation, given the known observation information, locations to be interpolated may change according to actual needs.
To output consistent results for a certain location, it should be independent of the information from other unobserved locations.
\end{itemize}

In this work, we propose a \textbf{S}elf-supervised learning framework to improve rainfall \textbf{S}patial \textbf{IN}terpolation by mining latent spatial patterns in historical observation data, which is  called \textbf{SSIN}.
To address the above issues, we design the SpaFormer (\textbf{Spa}tial Trans\textbf{Former}) model as the core component of SSIN. 
Similar to BERT, SpaFormer stacks multiple Transformer encoders ~\cite{ml11} to model spatial correlations, fuse spatial rainfall information, and generate effective data representations for target locations.
Specifically, three major techniques are employed to extend SpaFormer to be an effective spatial interpolator:
\textbf{(A)} inspired by the relative position embedding (RPE)~\cite{ml6}, we propose a natural extension of attention mechanism by incorporating spatial relative position embeddings (SRPE), thus enabling it to handle spatial data;
\textbf{(B)} to generate embeddings for the numerical inputs, we propose to utilize the fully connected network (FCN) as a more flexible linear embedding layer;
\textbf{(C)} we devise a shielded self-attention mechanism to avoid aggregating information from unobserved locations.
By adopting a mask-and-recover task with rich self-supervised signals, SSIN enables the SpaFormer model to be an effective spatial interpolator: given an instance with known rainfall observations, for any location queries, SpaFormer can infer their rainfall values simultaneously.
Our \textbf{main contributions} are summarized as follows:
\begin{itemize}[leftmargin=*]
\item
We identify the limitations of pre-settings in existing works for capturing spatial correlations and propose a self-supervised learning framework SSIN to solve rainfall spatial interpolation.

\item
We design a novel SpaFormer model as the core component of SSIN to overcome the shortcoming of existing methods. SpaFormer can learn informative embeddings for raw data, then adaptively model interactions and aggregate spatial context information for interpolation, instead of relying on any prior knowledge to characterize spatial correlations.

\item
We conduct extensive experiments on two real-world raingauge datasets, and the results demonstrate the effectiveness of our proposed method: on the HK and BW datasets, the RMSE is reduced by 12.28\% and 5.67\%, and the MAE is reduced by 6.97\% and 6.18\%, respectively.

\item
We take traffic spatial interpolation as another use case and conduct additional experiments, and the results further confirm the effectiveness and generality of our proposed method.

\end{itemize}

The remainder of this paper is organized as follows. 
Section~\ref{sec:Related_Work} introduces the related work.
In Section~\ref{sec:Methodology}, we elaborate on our methodology.
We analyze the experimental results in Section~\ref{sec:Experiments}. 
Finally, we conclude our work in Section~\ref{sec:Conclusions}.

%% file: 2-Related_work.tex
\section{Related Work} \label{sec:Related_Work}

\textbf{Rainfall Spatial Interpolation.}
Rainfall spatial interpolation is a widely studied task in Environmental Science, Geography, Water Science, and so on. Traditional spatial interpolation algorithms can be divided into two main categories~\cite{hy3,hy13}: (1) deterministic methods; (2) geostatistical methods.
Deterministic methods are directly based on surrounding measurements or specified formulas, while geostatistical approaches utilize empirical semivariograms to describe spatial correlations.
Deterministic methods include Inverse Distance Weighting (IDW)~\cite{hy5}, Triangular Irregular Network (TIN)~\cite{hy4}, Spline~\cite{hy10} and so on. 
Kriging is a generic name for a number of geostatistical techniques~\cite{hy8}. Ordinary Kriging (OK)~\cite{hy6} is the basic form, while other variants like Universal Kriging (UK)~\cite{hy9} incorporate additional variables on the basis of OK. 
In the fields of data mining and machine learning, rainfall spatial interpolation is still under-explored. Recent works mainly focus on developing GNNs-based solutions for spatial interpolation tasks, such as KCN~\cite{ml5} and IGNNK~\cite{ml16}. 
However, they are not specialized in handling rainfall spatial interpolation and suffer from one or two of the following drawbacks: (1) assume the existence of node attributes; (2) rely on the fixed adjacency matrix and ignore the fact that various rainfall events may have different spatial correlations.
Different from these studies, we fully consider the challenges of real-world rainfall interpolation and propose solutions to adaptively capture the intrinsic correlations from the spatial rainfall data itself.

\noindent
\textbf{Self-Supervised Learning.}
Self-supervised learning is a popular paradigm with the ability to learn the intrinsic correlations from the data itself.
The general process of self-supervised learning is to first construct the training signals directly from the raw data, and then train the model using the predefined optimization objective~\cite{ml17}.
Recently, self-supervised learning with Transformer~\cite{ml1} has achieved great success in natural language processing (NLP).
The most prominent model BERT~\cite{ml9} and its proposed masked prediction task are advancing the development of other domains, such as computer vision~\cite{ml28,ml29}, recommender systems~\cite{ml17} and database systems~\cite{ml30,ml31}.
In this work, we aim to develop the Transformer architecture as a spatial model and utilize self-supervised learning to solve rainfall spatial interpolation.

%% file: 3-Methodology.tex
\section{Methodology} \label{sec:Methodology}

\begin{table}[t]
\caption{Primary notations.}
\centering
\resizebox{0.65\columnwidth}{!}{
\begin{tabular}{|c|l|}
\hline
\textbf{Notation} & \textbf{Description}                       \\ \hline
$p_i$              & The $i$-th location.                      \\ \hline
$x_i$            & The rainfall value of location $p_i$.   \\ \hline
$\boldsymbol{r}_{ij}$  & The relative position of $p_j$ to $p_i$. \\ \hline
$\mathbf{Q}, \mathbf{K}, \mathbf{V}$     & Queries, Keys, Values embedding matrix.  \\ \hline
$\boldsymbol{q}_i$, $\boldsymbol{k}_i$, $\boldsymbol{v}_i$ & The $i$-row of $\mathbf{Q}, \mathbf{K}, \mathbf{V}$ \\ \hline
$d_e$, $d_k$, $d_{ff}$              & The hidden dimensions.       \\ \hline
$\boldsymbol{e}_{i}$              & The embedding vector for rainfall value $x_i$.  \\ \hline
$\boldsymbol{c}_{ij}$            & The embedding vector for relative position $\boldsymbol{r}_{ij}$.  \\ \hline
$\boldsymbol{W}^{(i)}$, $\boldsymbol{b}^{(i)}$ & The learnable matrix and bias for $i$-th FCN layer. \\ \hline
\end{tabular}
}
\label{tab:notation}
\end{table}

We first give the problem statement, then provide our SSIN framework, and finally introduce the detailed designs of SpaFormer.
The frequently-used notations\footnote{Spatial locations need to be rearranged into a sequence as model input, we may refer to a ``location'' as a ``node'' when describing it in the model.} in the paper are listed in Table~\ref{tab:notation}.

\subsection{Problem Statement}
In this paper, we focus on rainfall spatial interpolation task.
Assume there are $m$ rainfall monitoring stations, the known spatial rainfall information can be represented as $\left\{\langle p_i, x_i \rangle \right\}_{i=1}^{m}$, where $p_i$ denotes a location, $x_i$ is the rainfall value of $p_i$. 
Given an arbitrary location $p_k$, rainfall spatial interpolation is to estimate its rainfall value $\hat{x}_k$ according to $\left\{\langle p_i, x_i \rangle \right\}_{i=1}^{m}$.

\begin{figure*}[t]
\centering
\includegraphics[width=0.98\textwidth]{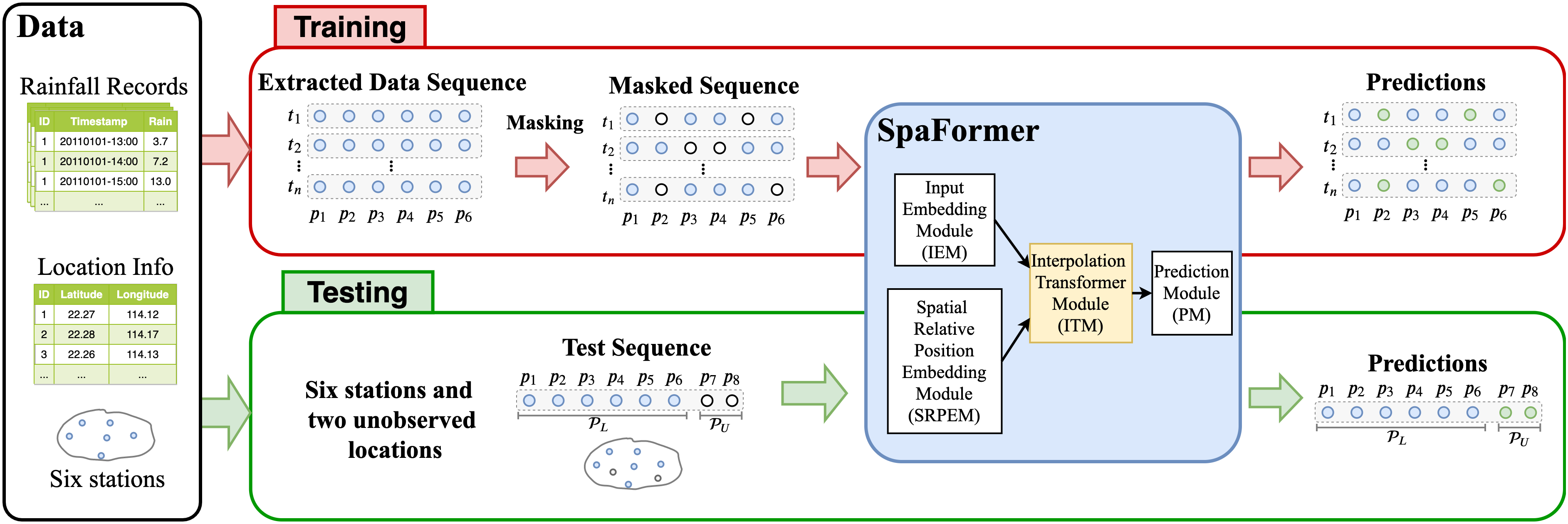}
\caption{The SSIN framework. One example with six stations and two locations to be interpolated is given.}
\label{fig:framework}
\end{figure*}

\subsection{SSIN Framework} \label{sec:framework}
Figure~\ref{fig:framework} shows an overview of the SSIN framework.

\noindent
\textbf{Data.}
In the climate database, each rainfall record usually consists of the station ID, timestamp, and rainfall value; the station information includes ID, longitude, and latitude.

\noindent
\textbf{Training.} SSIN first arranges the stations into a sequence and calculates the relative position information (including distance and azimuth, which will be discussed in Section \ref{sec:SRPEM}) of all location pairs.
Then SSIN extracts rainfall data based on the station sequence from the database and performs random masking and data standardization to generate the training sequences. 
Next, SSIN feeds the training sequences and relative position information into the SpaFormer model, and trains the model to accurately recover the rainfall values of masked locations.
\begin{itemize}[leftmargin=*]
\item
\textbf{Masking Strategy.}
We adopt a mask-and-recover training style like Masked Language Model (MLM).
During training, we randomly mask a portion of nodes in each sequence and train the model to predict the rainfall values of masked nodes.
Different from MLM, no special token (like $[MASK]$) is used since a separate embedding may negatively impact the embedding learning for continuous rainfall values.
Instead, we use the mean value of known nodes in the sequence to replace the input values of masked nodes\footnote{The same filling strategy is also used in the testing stage, that is, use the mean of observations to replace the input values of unobserved nodes.}.
For the spatial interpolation task, the mean value is more informative than zero because it can help masked (or unobserved) nodes directly obtain the average rainfall information.
To generate richer training signals, we adopt the dynamic masking strategy proposed in~\cite{ml15} --- the masking pattern is generated every time when each sequence is fed to the model.

\item
\textbf{Data Standardization.}
To accelerate the model convergence, we standardize the input data before feeding them into the model.
Considering that spatial rainfall may vary greatly at different time, we implement an instance-wise standardization for rainfall observations: the rainfall values $x_i$ at time $t$ is standardized using the mean and standard deviation of the known observed values $X_L$ at time $t$.
But the spatial position information is static and relative positions between all locations are within a certain range. 
Hence, we adopt a global standardization for the relative positions; that is, all distances and azimuths are respectively standardized using the statistics of the known locations.

\item
\textbf{Optimization.}
We use mean squared error as the objective function to calculate the loss:
$\mathcal{L} = \frac{1}{N} \sum_{i=1}^{N}\left(y_i-\hat{y}_i\right)^{2}$,
where $y_i$ is the true label and $\hat{y}_i$ is the predicted label in Prediction Module.
Instead of reconstructing the whole input, only masked nodes are predicted and used to calculate the loss.
\end{itemize}

\noindent
\textbf{Testing.}
Given rainfall data from monitoring stations at an arbitrary time, SSIN can make use of the trained SpaFormer model to infer the rainfall values for any locations without observations.

\subsection{SpaFormer Model} \label{sec:SpaFormer}
SpaFormer includes four modules: Input Embedding Module (IEM), Spatial Relative Position Embedding Module (SRPEM), Interpolation Transformer Module (ITM), and Prediction Module (PM).

\subsubsection{Input Embedding Module (IEM)}
This module takes the observed value $x_i$ as input and generates its embedding vector $\boldsymbol{e}_{i}$.

The input embedding module in language models is implemented by assigning a unique embedding to each token.
However, such an embedding strategy is not applicable in our setting, since rainfall observations are recorded in numerical values and it is not feasible to assign each possible value with a unique embedding.
The common embedding method for numerical input is to use an embedding vector (i.e., a linear embedding without bias) to map the input value to the latent embedding space~\cite{ml18, ml19}, as follows:
\begin{gather}
\boldsymbol{e}_{i}= x_i \cdot \boldsymbol{g}
\end{gather}
where $\boldsymbol{g} \in \mathbb{R}^{d}$ is the learned vector for embedding and $x_i$ is a scalar value.
However, this simple approach leads to two problems.
First, the representation capacity is limited since there is just a linear scaling relationship between all embeddings. 
Second, a zero input will be mapped to a zero embedding vector which will cause zero interactions in the latter self-attention computation. 
In the rainfall field, a standardized zero value is valid information to represent the average of current spatial rainfall and should not be ignored.

To tackle the above issues, we adopt a more flexible linear transformation to generate embeddings for the numerical inputs. That is, we transform $x_i$ into an embedding vector using a two-layer fully connected network (FCN) with hidden units $[d_{e}, d_{e}]$:
\begin{gather}
\boldsymbol{e}_{i}=\left(x_i \boldsymbol{v}_{x}^{(1)} + \boldsymbol{b}_{x}^{(1)}\right) \boldsymbol{W}_{x}^{(2)} +\boldsymbol{b}_{x}^{(2)} \in \mathbb{R}^{d_e}
\end{gather}
where $\boldsymbol{v}_{x}^{(i)}$ is the learnable vector, $\boldsymbol{W}_{x}^{(i)}$ the learnable matrix and $\boldsymbol{b}_{x}^{(i)}$ is the learnable bias vector for the $i$-th layer, $d_{e}$ is the hidden dimension.
The two-layer FCN introduces more parameters to improve the capacity and expressiveness of representation, and the existence of bias avoids the zero-embedding problem of zero values.

\subsubsection{Spatial Relative Position Embedding Module (SRPEM)} \label{sec:SRPEM}
This module generates the embedding vector $\boldsymbol{c}_{ij}$ for the spatial relative position $\boldsymbol{r}_{ij}$ between points $p_i$ and $p_j$.

Since the self-attention itself is a position-agnostic operation, how to explicitly encode position information is a crucial step in performing accurate interpolation.
The original Transformer adopts absolute position embedding (APE)~\cite{ml11}, in which the absolute position information is added to the token embeddings to serve as the model input.
However, absolute positions are less informative for our spatial interpolation task since hourly rainfall values have a weak correlation with the static longitudes and latitudes due to large spatiotemporal variations.
In fact, the existing work~\cite{ml34} has shown that the addition operation in APE will bring mixed and noisy correlations between two heterogeneous information resources.
Recent work~\cite{ml6} proposed relative position embedding (RPE), which incorporates relative position information into the self-attention mechanism.
Considering that the key point of spatial interpolation is to capture pair-wise correlations between locations, it is a natural way to employ RPE to encode position information.

\begin{figure}[t]
\centering
\includegraphics[width=0.7\columnwidth]{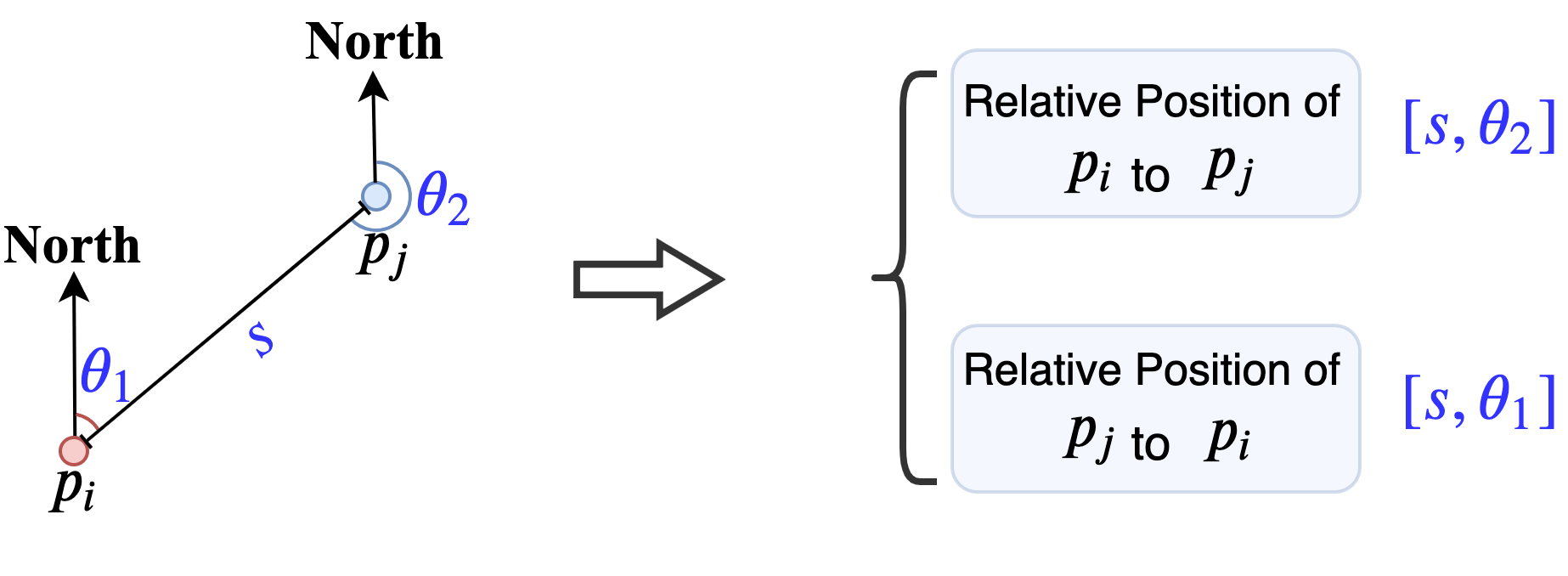}
\caption{Illustration of the spatial relative position.}
\label{relative_position}
\end{figure}

A look-up mechanism is not suitable to generate relative position embeddings for the spatial interpolation task, since spatial positions are recorded in real numbers and there are infinite possible pairs of relative positions.
To tackle this issue, we generate the spatial relative position embeddings (SRPE) by using a similar method to the Input Embedding Module.
We noticed that in addition to distance, the direction is also an important factor in describing relative positions between locations in 2D space.
Therefore, we use two values, distance and azimuth\footnote{Azimuth denotes the angle between the north direction and the line connecting the two locations.}, to represent the relative position. 
As shown in Figure~\ref{relative_position}, given a pair of locations $p_i$ and $p_j$, the relative position of $p_j$ to $p_i$ is $\boldsymbol{r}_{ij}=[s, \theta_1]$ while the relative position of $p_i$ to $p_j$ is $\boldsymbol{r}_{ji}=[s, \theta_2]$.
Then, we employ a two-layer FCN with hidden units $[d_e, d_e]$ to generate the embedding vector $\boldsymbol{c}_{ij}$:
\begin{gather}
\boldsymbol{c}_{ij}=\left(\boldsymbol{r}_{ij} \boldsymbol{W}_{r}^{(1)} + \boldsymbol{b}_{r}^{(1)}\right) \boldsymbol{W}_{r}^{(2)} +\boldsymbol{b}_{r}^{(2)} \in \mathbb{R}^{d_e}
\end{gather}
where $\boldsymbol{W}_{r}^{(i)}$ is the learnable matrix and $\boldsymbol{b}_{r}^{(i)}$ is the learnable bias for the $i$-th layer.

\subsubsection{Interpolation Transformer Module (ITM)} \label{sec:ITM}
The Interpolation Transformer Module is composed of a stack of identical layers and each layer has two components: a shielded self-attention with SRPE and a feed-forward network. Similar to the original implementation~\cite{ml11}, we employ a residual connection and a layer normalization for the two sub-layers, that is, the output of each sub-layer is $x = \text{LayerNorm}(x+\text{Sublayer}(x))$.

\begin{figure}[t]
\centering
\subfloat[Original Attention]{
\centering
\includegraphics[height=0.35\columnwidth]{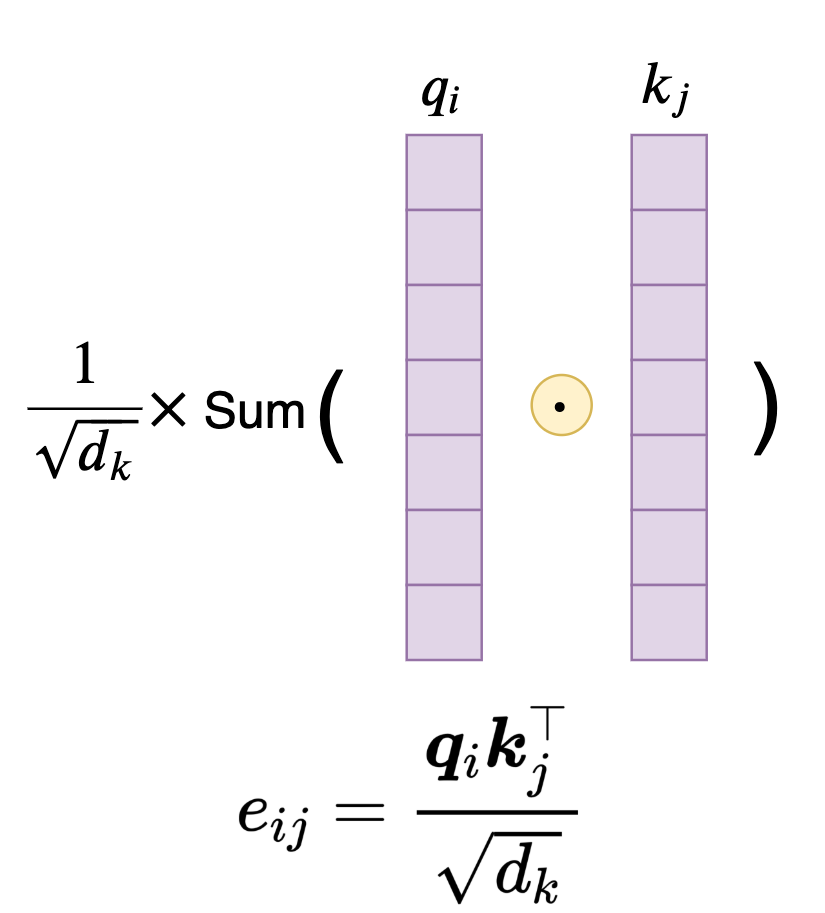}
\label{img:full_attn}
}\hspace{5mm}
\subfloat[Attention with SRPE]{
\centering
\includegraphics[height=0.35\columnwidth]{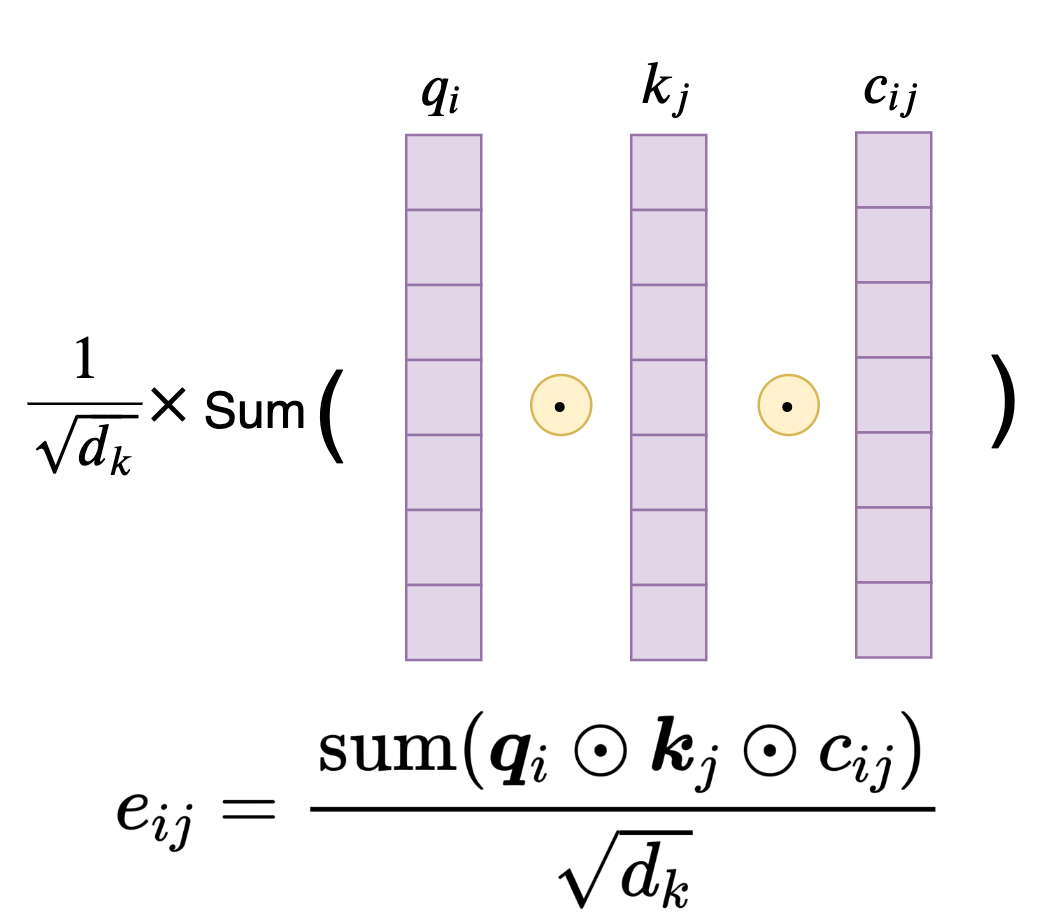}
\label{img:mask_attn}
}
\caption{Comparison between the original attention calculation and our attention calculation with inserting SRPE.}
\label{img:attn_SPRE}
\end{figure}

\textbf{Shielded Self-attention with SRPE.}
In the self-attention operation, we make two improvements to fit the spatial interpolation task: (\romannumeral1) use a natural extension method to incorporate the spatial relative position embedding (SRPE) for modeling pairwise relationships; (\romannumeral2) adopt the shielded mechanism to avoid aggregating information from unobserved nodes.
More specifically, let $\boldsymbol{E}=[\boldsymbol{e}_{1}, .., \boldsymbol{e}_{n}]$ denote the input embedding matrix for all nodes.
Then based on the input embeddings, we can calculate the queries $\boldsymbol{Q}=\boldsymbol{EW}^Q$, keys $\boldsymbol{K}=\boldsymbol{EW}^K$, values $\boldsymbol{V}=\boldsymbol{EW}^V$ by linear transformation, here $\boldsymbol{W}^Q$, $\boldsymbol{W}^K$, $\boldsymbol{W}^V$ are parameter matrices.
Let $\boldsymbol{q}_i$, $\boldsymbol{k}_i$, $\boldsymbol{v}_i$ $\in \mathbb{R}^{d_k}$ denote the query, key, and value embeddings of the $i$-th node, i.e., $i$-th row of $\boldsymbol{Q}$, $\boldsymbol{K}$, $\boldsymbol{V}$. 
Then in the attention, the output of the $i$-th node, $\boldsymbol{z}_i$ is calculated as below:
\begin{gather}
\label{eq:zi}
\boldsymbol{z}_{i}=\sum_{j=1}^{n} \alpha_{ij} \boldsymbol{v}_{j} \in \mathbb{R}^{d_k} \\
\alpha_{i j}=\frac{\exp(e_{ij})}{\sum_{k=1}^{n} \exp(e_{ik})} \\
\label{eq:eij}
e_{ij}=\frac{\text{sum}\left( \boldsymbol{q}_{i} \odot \boldsymbol{k}_{j} \odot \boldsymbol{c}_{ij} \right)}{\sqrt{d_k}}
\end{gather}
where $\alpha_{i j}$ is the normalized weight by using a softmax function, $e_{ij}$ is the attention score from node $j$ to $i$.
For the calculation of Equation (\ref{eq:eij}), we simply set $d_k=d_e$ and use the sum over the element-wise product of $\boldsymbol{q}_i$, $\boldsymbol{k}_i$ and $\boldsymbol{c}_{ij}$ to insert SRPE into the attention. 
As shown in Figure~\ref{img:attn_SPRE}, such an attention calculation with SRPE is a natural extension of the original attention.

Here, $\boldsymbol{z}_i$ is a spatial context-aware representation for location $p_i$, which is calculated by combining observation information and spatial relative position information.
To achieve a more flexible way of capturing complex spatial correlations, we adopt multiple heads to create different $\boldsymbol{Q}$/$\boldsymbol{K}$/$\boldsymbol{V}$ and learn distinct interactions separately.
Then we concatenate the outputs from different heads and use a projection to generate the final representation $\boldsymbol{z}_i^*$:
\begin{gather}
\boldsymbol{z}_i^* = \left(\boldsymbol{z}_i^{(1)} \| ... \| \boldsymbol{z}_i^{(h)} \right) \boldsymbol{W}^O \in \mathbb{R}^{d_e}
\end{gather}
where $\boldsymbol{z}_i^{(h)}$ is the output of the $h$-th head, $\boldsymbol{W}^O$ is the parameter matrix.

The pre-trained models like BERT simply adopt self-attention for language modelling, which is essentially full self-attention.
As shown in Figure~\ref{img:full_attn}, in full self-attention, each node aggregates information from all nodes without distinguishing them.
However, such full attention is not suitable to learn informative representations for the spatial interpolation task.
One reason is that the locations to be interpolated may vary according to the real needs, aggregating the information from other unobserved locations will make the model produce inconsistent results for a certain location.
To tackle this problem, we propose \textbf{Shielded Attention} (as shown in Figure~\ref{img:mask_attn}) to cut off the connections between each query node and other unobserved nodes:
(1) each observed node aggregates information from all observed nodes;
(2) each unobserved node aggregates information from itself and all observed nodes.
Here, we give an example to state the motivation and role of the shielded attention.
Given a network of rain gauges, at a given time, queries may involve different subsets of locations to be interpolated, e.g., locations $\{a, b, c\}$ and $\{c, d, e\}$. If adopting the full self-attention, the interpolated results of location $c$ in answering two queries will be inconsistent since they will aggregate information from locations $\{a, b\}$ and $\{d, e\}$, respectively. By using the shielded self-attention, the interpolated results can be prevented from being affected by other unobserved nodes\footnote{It is worth mentioning that in the original Transformer~\cite{ml1}, authors adopted the masked self-attention to avoid future information leakage in the decoding phase. Here, the shielded self-attention is proposed to prevent information of unobserved locations from interfering with interpolation results.}.
Besides, our results in Section~\ref{sec:ablation} show that the shielded mechanism can also help improve interpolation performance, since avoiding aggregating noisy information of unobserved nodes can learn better spatial context-aware representations\footnote{Unobserved locations do not own real rainfall values, their initialized values are likely to be inconsistent with the real rainfall field, hence bringing the noise to other nodes.}.

\begin{figure}[t]
\centering
\subfloat[Full Attention]{
\centering
\includegraphics[width=0.4\columnwidth]{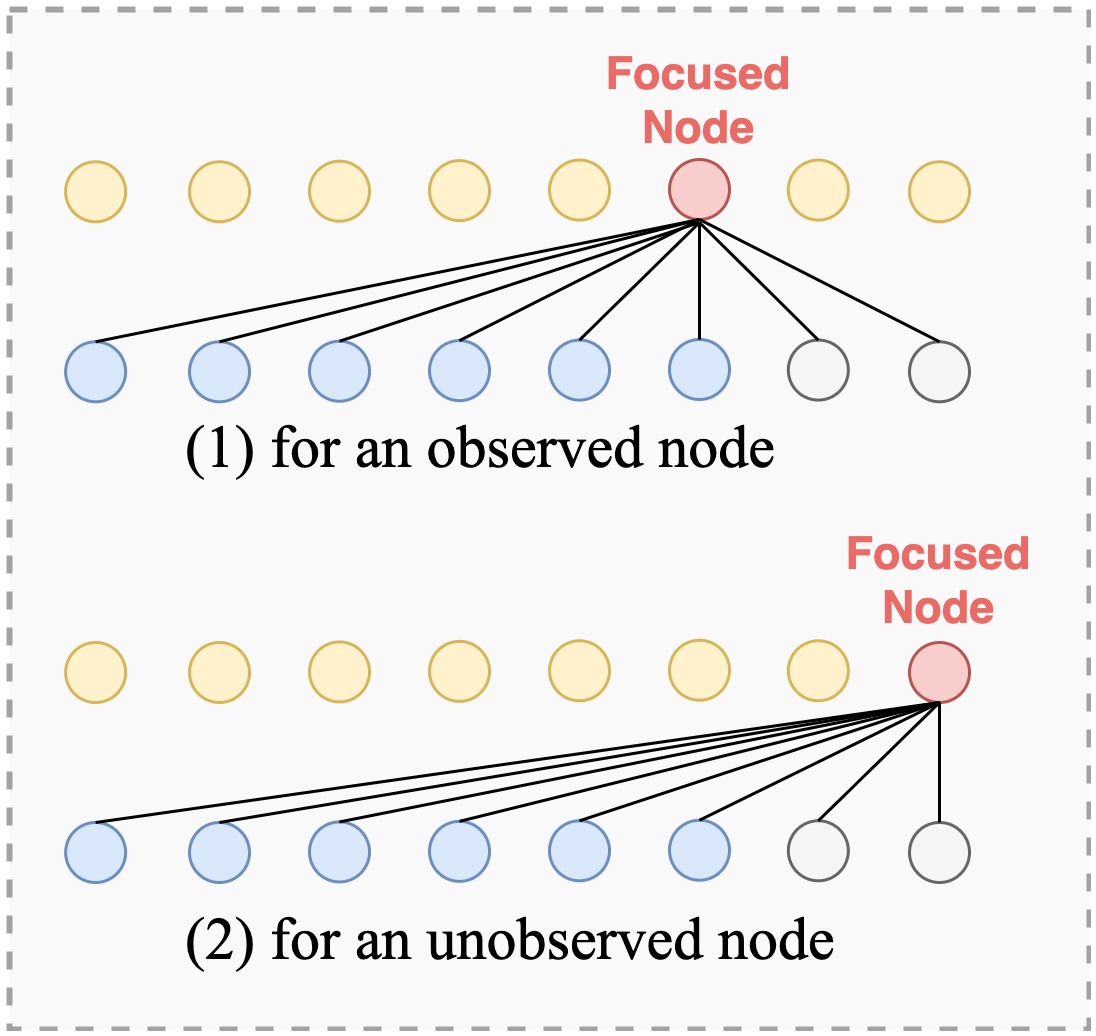}
\label{img:full_attn}
}\hspace{4mm}
\subfloat[Shielded Attention]{
\centering
\includegraphics[width=0.4\columnwidth]{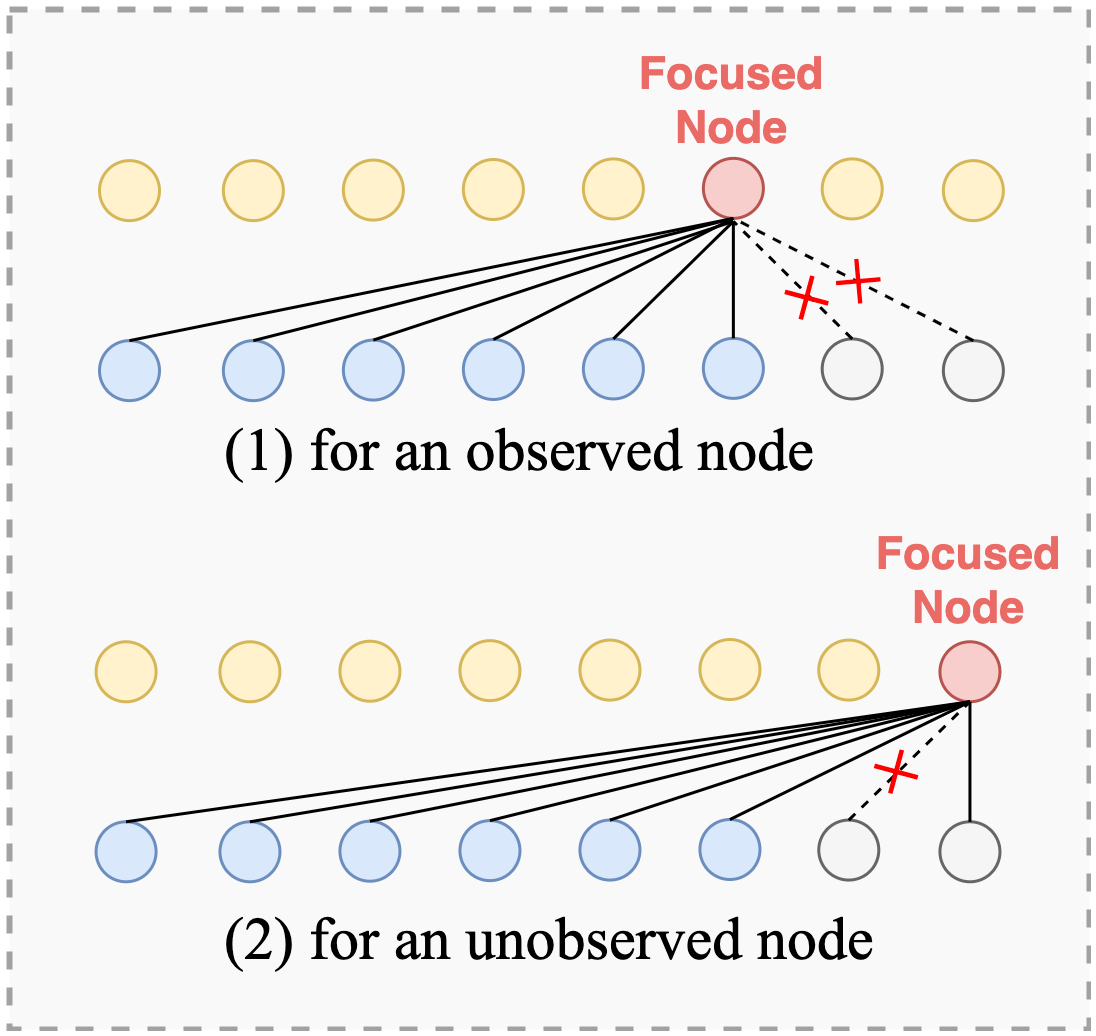}
\label{img:mask_attn}
}
\caption{Illustration of full attention and shielded attention on a focused node. For simplicity, only the connections of the focused node (in \textcolor[RGB]{234,107,102}{Red}) are displayed.
\textcolor[RGB]{108,142,191}{Blue} color means the input nodes with real observed rainfall, \textcolor{gray}{Gray} color denotes the input nodes with no observations. }
\label{img:attn}
\end{figure}

Since a sparse attention mechanism is not supported in existing deep learning libraries like PyTorch and TensorFlow,
a na\"ive implementation of Shielded Attention is first to calculate the attention scores between all pairs and then mask out the illegal connections.
However, this implementation is too time-consuming (the number of connections in attention is $O(L^2)$, $L$ is the sequence length).
Besides, the implementation with incorporating SRPE (i.e., Equation (\ref{eq:eij})) requires additional memories in a normal matrix operation due to inconsistent dimensions of matrices that store the queries, keys, and SRPEs.
To reduce the time and memory cost , we build a customized CUDA kernel by using TVM~\cite{ml14} to achieve the shielded attention with SRPE.
In Section~\ref{sec:Complexity}, we will given more details about the time and space complexity analysis. 

\textbf{Feed-Forward Network.}
The feed-forward network following the attention mechanism consists of a two-layer FCN with hidden units $[d_f, d_e]$, 
and a non-linear activation function (i.e., ReLU) is used between two layers:
\begin{gather}
\boldsymbol{h}_{i}=\operatorname{ReLU} \left(\boldsymbol{z}_i^* \boldsymbol{W}_{f}^{(1)} + \boldsymbol{b}_{f}^{(1)}\right) \boldsymbol{W}_{f}^{(2)} +\boldsymbol{b}_{f}^{(2)} \in \mathbb{R}^{d_e}
\end{gather}
where $\boldsymbol{W}_{f}^{(i)}$ is the learnable matrix and $\boldsymbol{b}_{f}^{(i)}$ is the learnable bias.

\subsubsection{Prediction Module (PM)}
Based on the node representation $\boldsymbol{h}_i$ from the Interpolation Transformer Module, we further derive the final estimated result via a two-layer FCN with hidden units $[d_e, 1]$, which is a regression task:
\begin{align}
\hat{y}_i =\left(\boldsymbol{h}_i \boldsymbol{W}_{p}^{(1)}  + \boldsymbol{b}_{p}^{(1)}\right) \boldsymbol{W}_{p}^{(2)} +\boldsymbol{b}_{p}^{(2)} \in \mathbb{R}
\end{align}
where $\boldsymbol{W}_{p}^{(i)}$ is the learnable matrix and $\boldsymbol{b}_{p}^{(i)}$ is the learnable bias.

\subsection{Discussion} \label{sec:Continuity}
\subsubsection{Numerical Embedding.}
Unlike NLP tasks that assign an individual (learnable) semantic embedding vector to each discrete element, the representation learning in spatial interpolation is depicted by the mapping function from the continuous original space to the continuous embedding space.
Specifically, there are two types of data representation to learn: the embedding mapping for numerical observations and the embedding mapping for spatial relative positions.
In this study, we adopt FCNs to generate embeddings for observations and spatial relative positions.
As a linear embedding mechanism, FCNs can learn consecutive embedding vectors for continuous inputs and the close inputs will be mapped into similar embeddings.
Instead of utilizing any pre-settings to capture spatial correlations, SpaFormer can learn the informative embeddings for numerical observations and relative spatial position information from historical data, then adaptively model the interactions of locations, and aggregate information to generate the spatial context-aware representation for target locations, thus estimating the rainfall values accurately.

\subsubsection{Complexity Analysis.}  \label{sec:Complexity}
As mentioned in Section~\ref{sec:ITM}, we build a customized CUDA kernel based on TVM to implement the shielded self-attention with SRPE.
In this section, we mainly analyze the role of such a customized CUDA kernel in reducing time and memory cost.
The empirical results about the memory and speed consumption will be shown in Section~\ref{sec:mem_time_cost}.

\noindent
\textbf{Time Complexity.}
The na\"ive implementation of shielded attention is essentially full connection, of which the number of Q-K pairs is $O(L^2)$.
For the attention operation with incorporating SPRE (i.e., Equation (\ref{eq:eij})), the computation complexity is $O(d_k)$. Hence, the na\"ive implementation takes time $O(L^2d_k)$.
In TVM implementation of shielded attention, each query node has at most $m+1$ valid connections, where $m$ is the number of observed nodes ($m<L$); for a sequence of length $L$, the all number of Q-K pairs is less than $(m+1)L$.
Hence, the TVM implementation can theoretically achieve a linear complexity $O(mLd_k)$ with regard to the sequence length.

\noindent
\textbf{Space Complexity.}
For one batch including $B$ sequences (of length $L$) in $H$-head attention, the shape of matrices that store the queries and keys are both $[B, H, L, d_k]$; during the training, all locations are known, all sequences can share one matrix of SRPEs and its shape is $[L, L, d_k]$.
To calculate the attention with SPRE (i.e, Equation (\ref{eq:zi})-(\ref{eq:eij})),
the normal matrix operation needs to extend the dimension of queries and keys to be $[B, H, L, L, d_k]$, so it will take memory $O(2BHL^2d_k+L^2d_k)$.
The TVM implementation can perform the calculation directly without dimension extension, so it takes memory $O(2BHLd_k+L^2d_k)$.

%% file: 4-Experiments.tex
\section{Experiments} \label{sec:Experiments}
\subsection{Experiment Setup}
\subsubsection{Datasets.}
We evaluate our proposed method on two real-world hourly raingauge datasets from different regions: Hong Kong (HK) in China and Baden-Württemberg (BW) in Germany. Due to the great variability of terrain, these two regions often suffer from regional rainfall-induced natural disasters, e.g., landslides~\cite{hy12} in Hong Kong and flash floods~\cite{hy11} in BW.

\begin{itemize}[leftmargin=*]
\item
\textbf{HK} raingauge data are obtained from the Hong Kong Observatory (HKO)\footnote{\url{https://www.hko.gov.hk/en/index.html}} and the Geotechnical Engineering Office (GEO)\footnote{\url{https://www.cedd.gov.hk/eng/about-us/organisation/geo/index.html}}. 123 rain gauges are available in this area, and data precision is 0.1-mm.
The rain hours between 2008 and 2012 are selected to be the final dataset with 3855 valid timestamps. 

\item
\textbf{BW} raingauge data are public data, which can be accessed in the Climate Data Center (CDC)\footnote{\url{https://www.dwd.de/EN/climate_environment/cdc/cdc_node_en.html}} of the German Weather Service (DWD). 
There are 132 rain gauges available in this area and the rainfall data are based on 0.1-mm precision.
This dataset spans from 2012 to 2014 and contains 3640 valid rainy hours.
\end{itemize}

For two datasets, we randomly sample 20\% rain gauges as the test locations, and the rest serves as the training data.
Table~\ref{tab:dataset} shows the dataset details. 

\begin{table}[h]
\caption{Dataset details.}
\resizebox{0.9\columnwidth}{!}{
\begin{tabular}{cccccc}
\hline
\multicolumn{1}{r}{Dataset} & Time Span & \#Rainy hours & \#Raingauges & \#Training nodes & \#Test nodes \\ \hline
HK                          & 2008-2012 & 3855         & 123          & 98               & 25              \\
BW                          & 2012-2014 & 3640         & 132          & 106              & 26              \\ \hline
\end{tabular}
}
\label{tab:dataset}
\end{table}

\subsubsection{Baselines.} \label{sec:baseline}
We compare SpaFormer with the following baseline methods.

\noindent
\textbf{(1) Traditional Interpolation Methods.} 
\begin{itemize}[leftmargin=*]
\item
\textbf{TIN:} Triangular Irregular Network~\cite{hy4}, a deterministic method that creates a series of triangles by using all sampled points, then interpolates with a weighted value of the apexes of the triangle.

\item
\textbf{IDW:} Inverse Distance Weighting~\cite{hy5}, a deterministic method that interpolates with a linear weighted sum of available points, the weights are calculated on a function of inverse distance.

\item
\textbf{TPS:} Thin Plate Spline~\cite{hy10}, a deterministic method that is a spline-based technique, in which the smoothing parameter is calculated by minimizing the generalized cross validation.

\item
\textbf{OK:} Ordinary Kriging~\cite{hy6}, a geostatistical interpolation method which assumes the stationarity of data and that the distance between locations reflect the spatial correlations.
\end{itemize}

\noindent
\textbf{(2) GNN-based Interpolation Methods.} 
\begin{itemize}[leftmargin=*]
\item
\textbf{KCN:} Kriging Convolutional Network~\cite{ml5}. The method constructs local subgraphs and predicts each center node’s label based on node features and neighboring labels. 

\item
\textbf{IGNNK:} Inductive Graph Neural Network Kriging~\cite{ml16}. The method treats time-series signals as the node features, generates random subgraphs, randomly mask some nodes, and reconstructs these signals.
\end{itemize}

\subsubsection{Metrics.}
To evaluate the performance of interpolation methods, we adopt RMSE (Root Mean Squared Error), MAE (Mean Absolute Error), and NSE (Nash–Sutcliffe’s Efficiency coefficient)~\cite{hy7} as evaluation metrics. 
$\mathrm{NSE}=1-\frac{\sum_{i=1}^{n}\left[y_{i}-\hat{y}_{i}\right]^{2}}{\sum_{i=1}^{n}\left[y_{i}-\bar{y}\right]^{2}}$ is a widely used indicator to assess the performance of hydrological models, where $\bar{y}$ is the mean value of all observed values.
NSE's value ranges from $-\infty$ to 1: the closer to 1, the better.

\subsubsection{Implementation Details.}
For traditional interpolation methods, we run different settings and report their best performance, where the power parameter of IDW is \textit{2}, the variogram model of OK is \textit{spherical}, and the type of TIN interpolating function is \textit{linear}; for TPS, there is no parameters need manual tuning.
For KCN and IGNNK, we use the public code provided by their authors.
To better validate their performance, we search for the best hyperparameters in much larger search space than that in the original papers.
The tuning parameters include learning rate, weight decay, dropout rate, hidden dimension, and the kernel length of the adjacency matrix.
Table~\ref{tab:hp_search} summarizes the value ranges of the hyperparameter search.
Other settings of KCN and IGNNK, like optimizer and activation function, are kept the same as original works.
For IGNNK, the time dimension is set as 1 to compare with other spatial interpolators.

We implement our method with PyTorch. We denote the number of Transformer blocks as $T$, the number of attention heads as $H$, the embedding dimension as $d_e$, the dimension of query/key/value as $d_k$, the hidden dimension of the feed-forward network as $d_f$.
In our implementation, $T=3$, $H=2$, $d_e=d_k=16$, $d_f=256$.
We use Adam~\cite{ml24} as the optimizer with $\beta_{1}=0.9$, $\beta_{2}=0.98$, and $\epsilon=10^{-9}$. The warmup strategy~\cite{ml11} is adopted to vary the learning rate, and the warmup step is set as $1200$.
Batch size is set as 64.
We train our model for 100 epochs. 
For each epoch, we randomly mask 10 times for each sequence to generate different spatial patterns to augment the dataset, the mask ratio is set as 20\%.
All the experiments were run on a CentOS 7.9.2009 server equipped with a 72-core Intel(R) Xeon(R) Gold 6240 CPU and one Tesla V100 GPU.

\begin{table}[h]
\renewcommand\arraystretch{1.2}
\caption{Hyperparameter search space for KCN and IGNNK.}
\centering
  \begin{tabular}{p{3cm}p{4cm}<{\centering}}
    \toprule
    \normalsize Hyperparameter & \normalsize Range \\
    \midrule
    \small Learning rate  & \small (0, 0.01) \\
    \small Weight decay  & \small (0, 1e-3) \\
    \small Dropout rate & \small (0, 0.5) \\
    \small Hidden dimension & \small \{4, 8, 16, 32, 64, 128\}  \\
    \small Kernel length & \small \{10, 5, 1, 0.5, 0.1, 0.05, 0.01\} \\
    \bottomrule
  \end{tabular}
\label{tab:hp_search}
\end{table}

\subsection{Experiment Results}
\subsubsection{Overall Performance}  \label{sec:overall_result}
\begin{table}[t]
\caption{Overall performance of different methods. The best and the second best performance are denoted in bold and underlined fonts, respectively. For metrics, ``$\uparrow$'' means the higher the better while ``$\downarrow$'' means the lower the better.}
\resizebox{0.8\columnwidth}{!}{
\begin{tabular}{c|ccc|ccc}
\hline
\multirow{2}{*}{\textbf{Methods}} & \multicolumn{3}{c|}{\textbf{HK Dataset}}             & \multicolumn{3}{c}{\textbf{BW Dataset}}              \\ \cline{2-7} 
                                  & RMSE$\downarrow$ & MAE$\downarrow$ & NSE$\uparrow$   & RMSE$\downarrow$ & MAE$\downarrow$ & NSE$\uparrow$   \\ \hline
TIN                               & 3.0088           & 0.9684          & 0.7538          & 1.0985           & {\ul 0.3494}    & 0.4008          \\
IDW                               & 2.9171           & 1.1056          & 0.7686          & 1.0493           & 0.3917          & 0.4533          \\
TPS                               & {\ul 2.6594}     & {\ul 0.8953}    & {\ul 0.8076}    & 1.0985           & 0.3537          & 0.4008          \\
OK                                & 2.8661           & 1.0001          & 0.7766          & 1.0804           & 0.3647          & 0.4203          \\
KCN                               & 2.7122           & 0.9935          & 0.7999          & {\ul 1.0468}     & 0.3819          & {\ul 0.4559}    \\
IGNNK                             & 3.3007           & 2.0864          & 0.7037          & 1.1429           & 0.6018          & 0.3514          \\ \hline
SpaFormer                         & \textbf{2.3328}  & \textbf{0.8329} & \textbf{0.8520} & \textbf{0.9874}  & \textbf{0.3278} & \textbf{0.5158} \\
Improv.                           & 12.28\%           & 6.97\%           & 5.50\%           & 5.67\%            & 6.18\%           & 13.14\%          \\ \hline
\end{tabular}
}
\label{tab:main_result}
\end{table}

We evaluate our method and report the results in Table~\ref{tab:main_result}.
SpaFormer achieves the best performance on two hourly raingauge datasets.
Compared with other methods, SpaFormer does not require any prior knowledge to characterize the spatial correlations.
By constructing rich masking patterns as the training objective, SpaFormer learns the effective embeddings for numerical observations and spatial position information from historical data, then the spatial correlations are captured via self-attention that adaptively models the interactions of nodes.
From the results, we can see that such a purely data-driven method improves the interpolation performance by a large margin: reducing RMSE for the HK and BW datasets by 12.28\%, and 5.67\%.

Among all baselines, traditional interpolation methods even achieve better results than GNN-based solutions. 
Specifically, TPS is the best baseline on the HK dataset, while TIN obtains the second lowest MAE on the BW dataset. 
Two GNN-based solutions, KCN and IGNNK, cannot handle the hourly rainfall spatial interpolation well.
Even with the best hyperparameters searched in larger space, only KCN achieves slightly lower RMSE and MAE than the simple method IDW.
The main reason is that KCN and IGNNK rely on the pre-defined adjacency matrix to capture the spatial correlations and guide the message passing. However, the pre-defined adjacency matrix may not be optimal for rainfall interpolation all the time, thus limiting their performance.
Besides, their model architectures are flawed due to the lack of careful consideration of the characteristics of rainfall spatial interpolation.
For example, KCN constructs a subgraph with $K$ nearest neighbors around each center point and only predicts the value of the central point, which leads to a weak supervision signal; besides, the rainfall field is dynamically changing, and using a fixed-size subgraph may miss important distant neighbors.
IGNNK randomly masks some nodes to generate rich training signals; 
however, no specific design is proposed to prevent information of unobserved locations from interfering with interpolation results, which is verified to be important for accurate interpolation performance (see ablation study about the shielded mechanism in Section~\ref{sec:ablation}).

\begin{table}[t]
\caption{Model size and running time on two datasets.}
\resizebox{0.9\columnwidth}{!}{
\begin{tabular}{|c|cccc|cc|c|cl|}
\hline
\multirow{2}{*}{Dataset} & \multicolumn{4}{c|}{Model Size}                                                                         & \multicolumn{2}{c|}{Sequence} & Training                                                          & \multicolumn{2}{c|}{Inference}                                                          \\ \cline{2-10} 
                         & \multicolumn{1}{c|}{\#$T$} & \multicolumn{1}{c|}{\#$H$} & \multicolumn{1}{c|}{\#$d_k$} & \#Param                & Num          & Length         & \begin{tabular}[c]{@{}c@{}}Avg Time \\ per Epoch (s)\end{tabular} & \multicolumn{2}{c|}{\begin{tabular}[c]{@{}c@{}}Avg Time \\ per Seq (ms)\end{tabular}} \\ \hline
HK                       & \multirow{2}{*}{3}       & \multirow{2}{*}{2}       & \multirow{2}{*}{16}      & \multirow{2}{*}{33585} & 3855         & 123            & 19.5                                                              & \multicolumn{2}{c|}{2.6}                                                              \\
BW                       &                          &                          &                          &                        & 3640         & 132            & 19.2                                                              & \multicolumn{2}{c|}{2.7}                                                              \\ \hline
\end{tabular}}
\label{tab:size_time}
\end{table}

\begin{figure}[t]
\centering
\subfloat[Computation Time]{
\centering
\includegraphics[height=0.35\columnwidth]{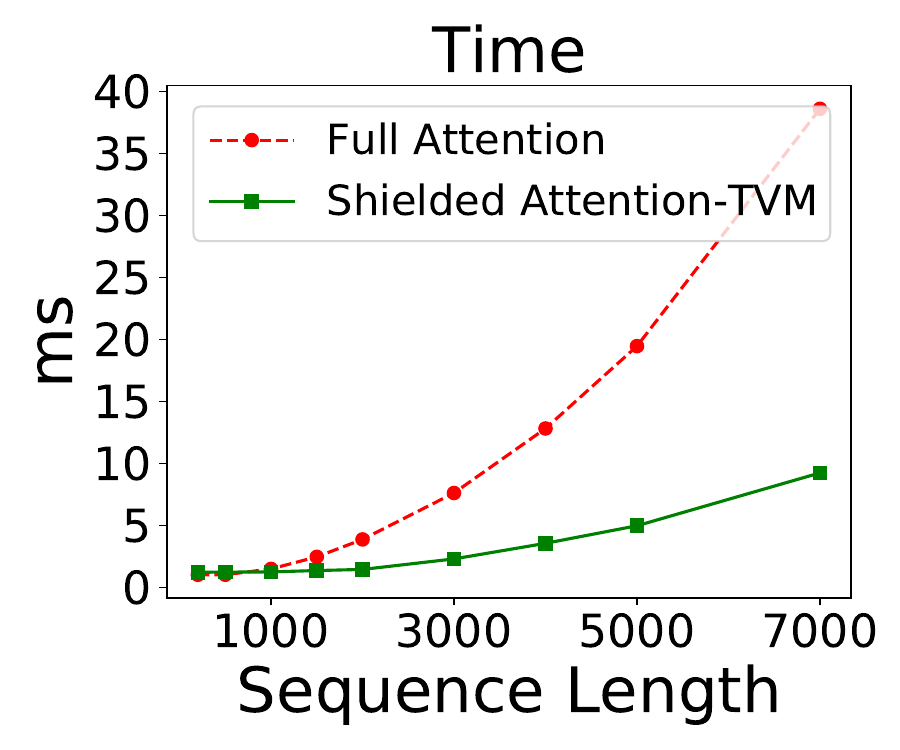}
\label{img:effi_time}
}\hspace{2mm}
\subfloat[Memory Occupation]{
\centering
\includegraphics[height=0.35\columnwidth]{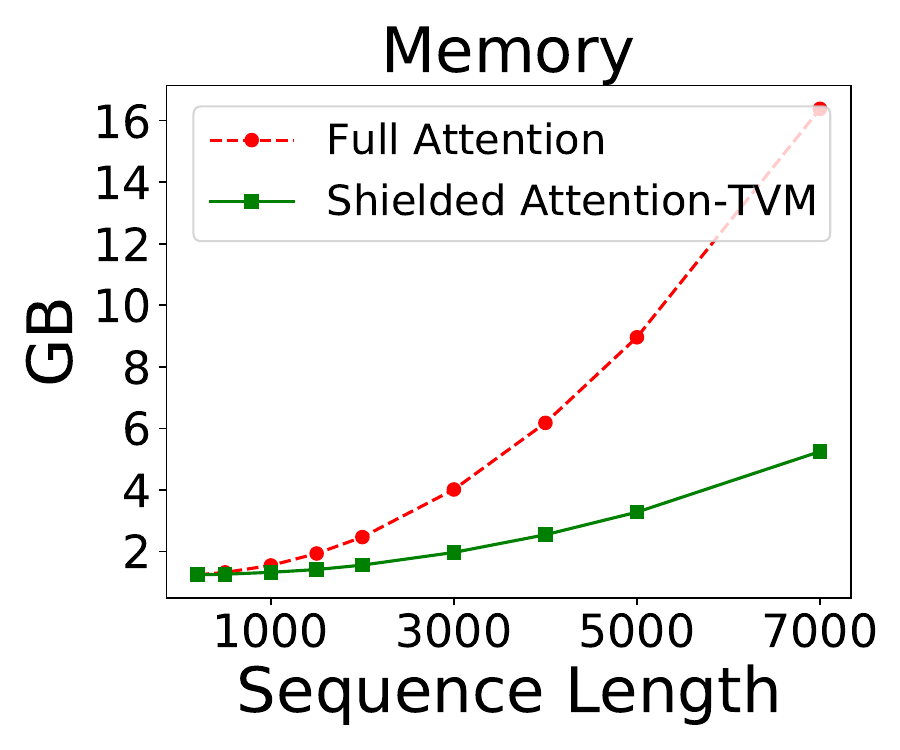}
\label{img:effi_memo}
}
\caption{Comparison of the time and memory consumption between the full attention and the TVM implementation of the shielded attention.}
\label{img:efficiency}
\end{figure}

\subsubsection{Memory and Speed Consumption} \label{sec:mem_time_cost}
We trained the SpaFormer model with $T=3$ blocks, $H=2$ heads and the hidden dimension $d_e=d_k=16$.
Table~\ref{tab:size_time} shows our model size and running time on two datasets.
We can see such a relatively small model can achieve a fast speed:
for HK and BW datasets, the average training time on each epoch is 19.5s and 19.2s, and the average inference time on each sequence is 2.6ms and 2.7ms.

Next, we show the efficiency of the customized TVM CUDA kernel for the shielded attention with SRPE.
Within a city or region, the number of stations is usually limited, which means the length of training sequences is relatively short.
In practical scenarios, researchers may need to interpolate thousands of locations to obtain fine-grained spatial rainfall information for an area. Performing interpolation for a lot of locations in parallel means a long testing sequence, in this case, the efficiency of the interpolation method will be an important issue.
In SpaFormer, we build a customized CUDA kernel implemented based on TVM to implement the shielded attention with SPRE.
Here, we compare the efficiency of full attention (i.e., the na\"ive implementation of shielded attention) with the TVM implementation of shielded attention. 
We take the situation of the HK region (i.e., 123 stations) as an example and show the comparison results.
Figure~\ref{img:efficiency} shows the time and memory cost with respect to the sequence length\footnote{The sequence includes observed and unobserved locations, hence the number of locations to be interpolated is $L-123$.} $L$.
. We can see that the time and memory cost of the TVM implementation can be much smaller than that of full attention, especially for long sequences.
When the sequence length reaches 7000 locations, full attention needs 38.6ms and 16.4GB while TVM implementation only requires 9.2ms and 5.2GB.
Our optimization with TVM is practical to use in real-world applications.

\subsubsection{Ablation Study} \label{sec:ablation}
In this part, we conduct an ablation study from two aspects to investigate the effectiveness of design choices in our work: the architecture of SpaFormer and the training strategy.
Firstly, we design six variants of our proposed SpaFormer: 
(1) ``emb: pos-l'' applies a linear layer without bias to generate embeddings for relative positions.
(2) ``emb: input-l'' ” applies a linear layer without bias to generate embeddings for input values.
(3) ``emb: both-l'' applies a linear layer without bias to generate embeddings for both input values and relative positions.
(4) ``attn: with SAPE'' adopts self-attention with spatial absolute position embedding; just like the original Transformer, we employ the addition operation to integrate the absolute position embeddings into the input embeddings as the model input; here, the absolute position embedding is generated from $[latitude, longitude]$ by using a two-layer FCN.
(5) ``attn: w/o shield'' applies the traditional self-attention without the shielded mechanism.
(6) ``na\"ive trans'' is a na\"ive version of Transformer architecture which applies a linear layer without bias as the embedding layers, uses self-attention with spatial absolute position embedding, and no shielded mechanism is applied.
Secondly, we design two variants to test the effects of the training strategy:
(7) ``static masking'' performs random masking for each sequence during data preprocessing and then trains the model using the generated masked data over multiple epochs.
(8) ``zero fill'' replaces the input values of masked/test nodes with zeros.

Table~\ref{tab:ablation} shows the results. 
First, compared with SpaFormer, ``emb: pos-l'' reduces performance slightly, followed by ``emb: input-l'' and finally ``emb: both-l''.
This verifies that: (1) using two-layer FCNs as the embedding layer is more expressive than a linear layer without bias; (2) without using bias, the zero-embedding issue in input embeddings will affect the normal interactions between nodes and significantly hurt performance.
Second, SpaFormer consistently outperforms both ``attn: with SAPE'' and ``attn: w/o shield'', which shows that: (1) relative position embedding is better at capturing the pairwise relationships and is more suitable for spatial interpolation; (2) the shielded mechanism is able to learn better spatial context-aware representations by avoiding aggregating noisy information from masked/unobserved nodes.
Third, the poor results of ``na\"ive trans'' confirm the effectiveness of the proposed techniques, and a na\"ive adaptation of the Transformer architecture is limited to perform accurate spatial interpolation due to the lack of careful consideration of the characteristics of spatial interpolation.
Finally, the results of ``static masking'' show rich masking patterns can help the model to yield a higher accuracy, the results of ``zero fill'' confirm that the average rainfall values are informative to learn better spatial context-aware representations for accurate interpolation.

\begin{table}[t]
\caption{Ablation study results.}
\resizebox{0.8\columnwidth}{!}{
\begin{tabular}{c|ccc|ccc}
\hline
\multirow{2}{*}{\textbf{Variants}} & \multicolumn{3}{c|}{\textbf{HK Dataset}}             & \multicolumn{3}{c}{\textbf{BW Dataset}}              \\ \cline{2-7} 
                                   & RMSE$\downarrow$ & MAE$\downarrow$ & NSE$\uparrow$   & RMSE$\downarrow$ & MAE$\downarrow$ & NSE$\uparrow$   \\ \hline
SpaFormer                          & \textbf{2.3328}  & \textbf{0.8329} & \textbf{0.8520} & \textbf{0.9874}  & \textbf{0.3278} & \textbf{0.5158} \\ \hline
emb: pos-l                         & 2.3417           & 0.8444          & 0.8505          & 1.0020           & 0.3451          & 0.5014          \\
emb: input-l                       & 2.7296           & 1.0237          & 0.7974          & 1.0779           & 0.3814          & 0.4231          \\
emb: both-l                        & 2.7846           & 1.0465          & 0.7891          & 1.1233           & 0.4413          & 0.3734          \\
attn: with SAPE                    & 2.4599           & 0.8999          & 0.8354          & 1.1149           & 0.3974          & 0.3828          \\
attn: w/o shield                   & 2.3868           & 0.8334          & 0.8451          & 1.2883           & 0.4415          & 0.1758          \\
naive trans                        & 3.7002           & 1.5344          & 0.6276          & 1.2225           & 0.4896          & 0.2579          \\ \hline
static masking                     & 2.3606           & 0.8462          & 0.8484          & 1.0080           & 0.3851          & 0.4955          \\
zero fill                          & 2.3945           & 0.8997          & 0.8441          & 1.0136           & 0.3718          & 0.4898          \\ \hline
\end{tabular}}
\label{tab:ablation}
\end{table}

\subsubsection{Parameter Sensitivity}
We perform a more detailed analysis to evaluate the effects of key hyperparameters in SpaFormer.

\begin{figure}[t]
\centering
\begin{minipage}[t]{0.6\textwidth}
\centering
\subfloat[HK Dataset]{
\centering
\includegraphics[width=0.45\columnwidth]{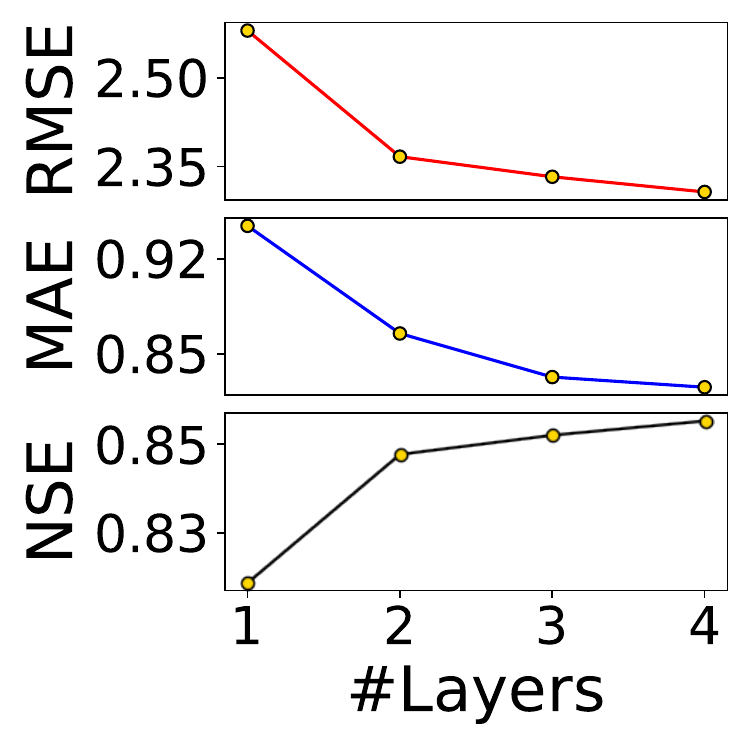}
\label{img:hk_layer_effect}
}\hspace{2mm}
\subfloat[BW Dataset]{
\centering
\includegraphics[width=0.45\columnwidth]{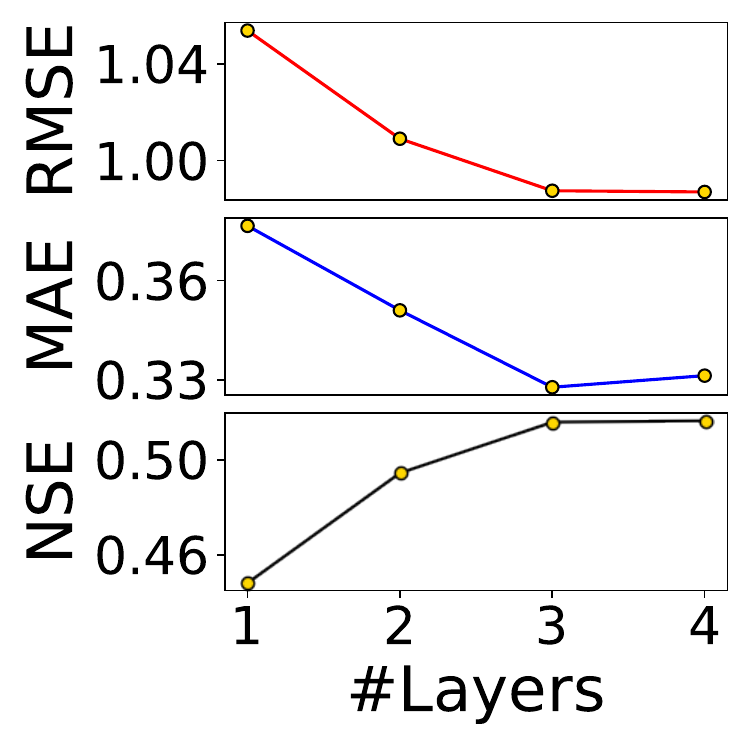}
\label{img:Bw_layer_effect}
}
\caption{Performance comparison w.r.t. the number of layers.}
\label{img:layer_effect}
\end{minipage}
\end{figure}

\textbf{Effect of Network Depth.}
Since the Interpolation Transformer Module stacks multiple identical layers, 
we are interested in how the performance changes w.r.t. the number of layers.
The results are summarized in Figure	~\ref{img:layer_effect}. 
We can see that the performance is poor when only one layer is used. 
As the number of layers increases, the performance of the model increases.
When the number of layers reaches three, the performance becomes relatively stable.
Considering a deeper model incurs more training overhead, we choose the configuration with three layers (i.e., $T=3$).

\begin{figure}[t]
\centering
\begin{minipage}[t]{0.65\textwidth}
\centering
\subfloat[HK Dataset]{
\centering
\includegraphics[width=0.45\columnwidth]{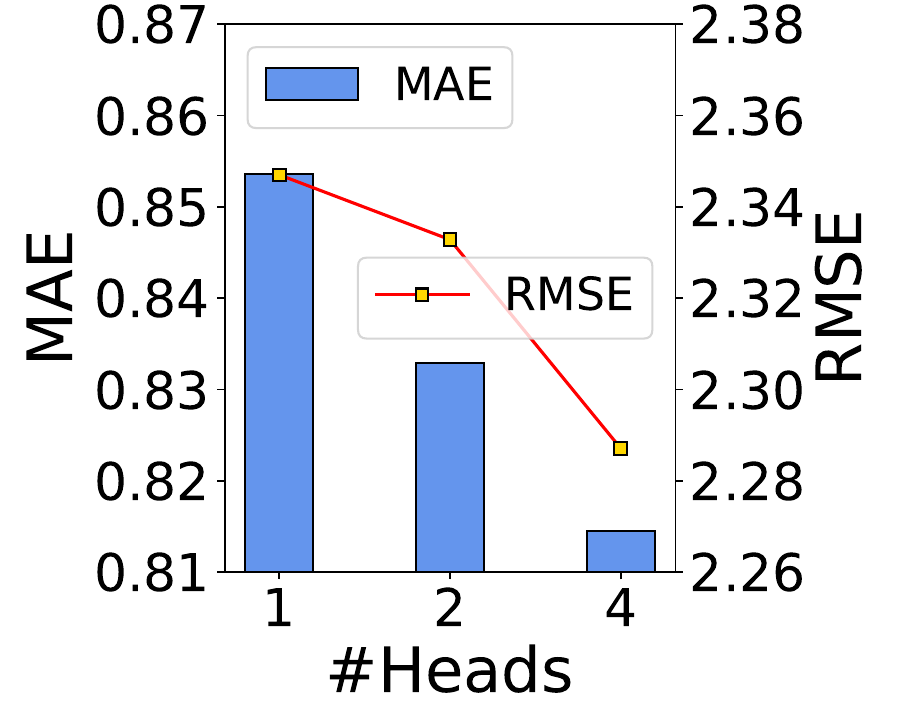}
\label{img:hk_head_effect}
}
\subfloat[BW Dataset]{
\centering
\includegraphics[width=0.45\columnwidth]{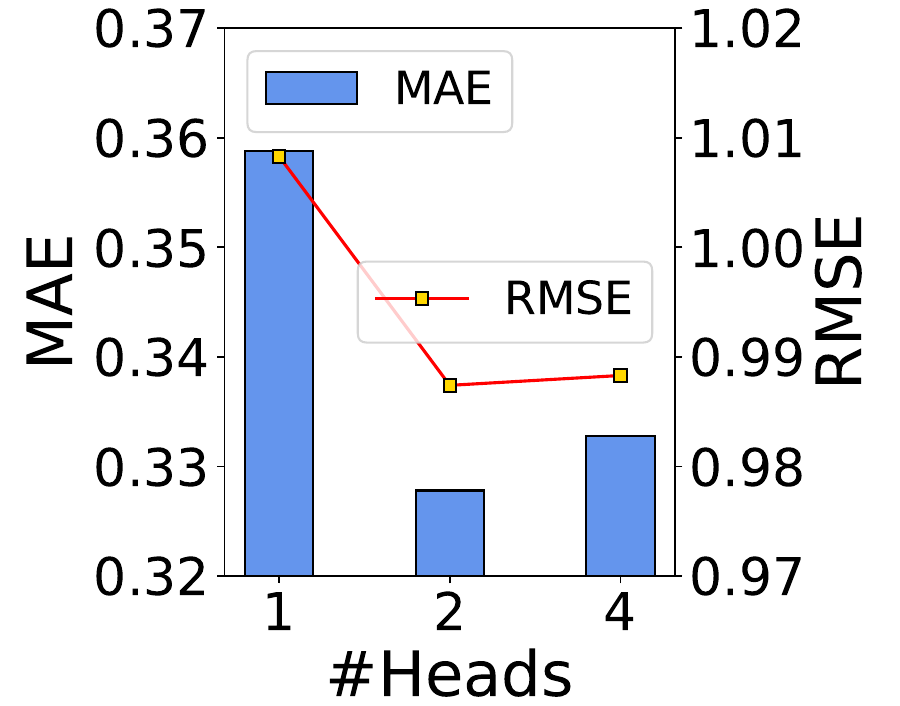}
\label{img:Bw_head_effect}
}
\caption{Performance comparison w.r.t. the number of attention heads.}
\label{img:head_effect}
\end{minipage}
\end{figure}

\textbf{Effect of the Number of Attention Heads.}
Then, we explore the effect of the number of attention heads.
The results are shown in	 Figure~\ref{img:head_effect}.
On the HK dataset, the performance continuously increases when we increase the number of attention heads. 
For the BW dataset, the best number of attention heads is 2.
In attention, the multi-head mechanism allows the model to jointly attend to information from different representation subspaces at different positions.
According to the results, we conjecture that the spatial distribution of HK rainfall data is more complicated than BW, thus requiring more attention heads to fit complex spatial relationships.

\begin{figure}[t]
\centering
\begin{minipage}[t]{0.6\textwidth}
\subfloat[HK Dataset]{
\centering
\includegraphics[width=0.45\columnwidth]{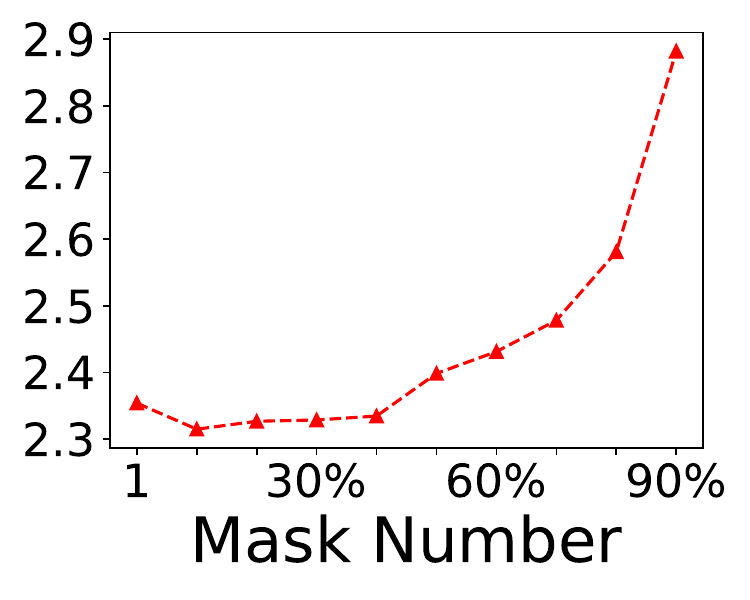}
\label{img:hk_mask_effect}
}\hspace{3mm}
\subfloat[BW Dataset]{
\centering
\includegraphics[width=0.45\columnwidth]{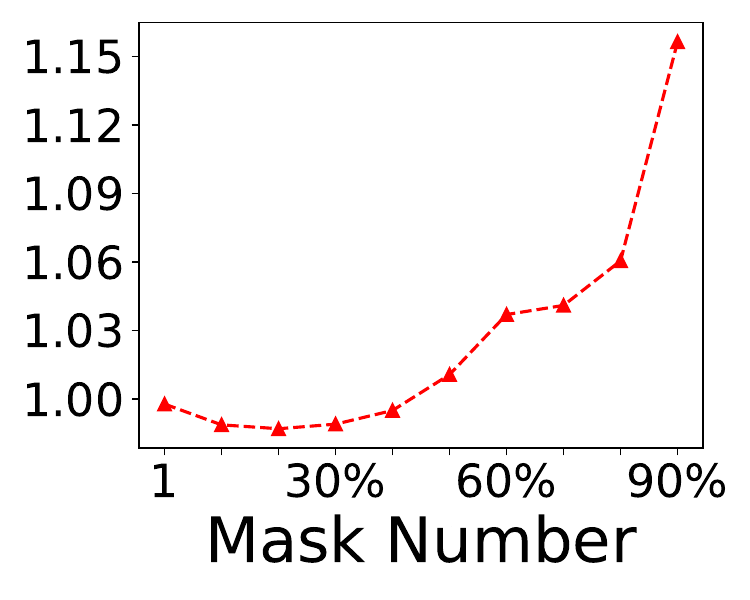}
\label{img:bw_mask_effect}
}
\caption{Performance (RMSE) comparison w.r.t. mask ratios.}
\label{img:mask_effect}
\end{minipage}
\end{figure}

\textbf{Effect of Mask Ratio.}
During the training data generation, the mask ratio is also an important hyperparameter. 
Here, we study the performance w.r.t. mask ratios, where we choose the mask number $l_m$ from $10\%$ to $90\%$ of the sequence length, plus with $l_m=1$ (i.e., the extreme case, only one node is masked).
As shown in Figure~\ref{img:mask_effect}, we can observe that the error generally decreases first and then increases as the mask number increases.
The results mean that too few masks are not enough to generate rich training signals, while too many masks make the input information insufficient for reliable training.
A mask ratio between $10\%$ and $30\%$ is a good balance.

\subsubsection{Training Data Amount and Model Update}
A large amount of rainfall record data implies rich spatial pattern information, thus enlarging the training data size is likely to improve the performance of SpaFormer.
We use more historical rainfall data to augment the training dataset
\footnote{Considering that very old data may have poor quality control, we mainly selected data after 2000.},
i.e., the double and triple size of the original data, then compare the performance. 
the results in Table~\ref{tab:data_amount} show that the performance continuously improves when more training data is used.

\begin{figure}[t]
\centering
\includegraphics[width=0.7\columnwidth]{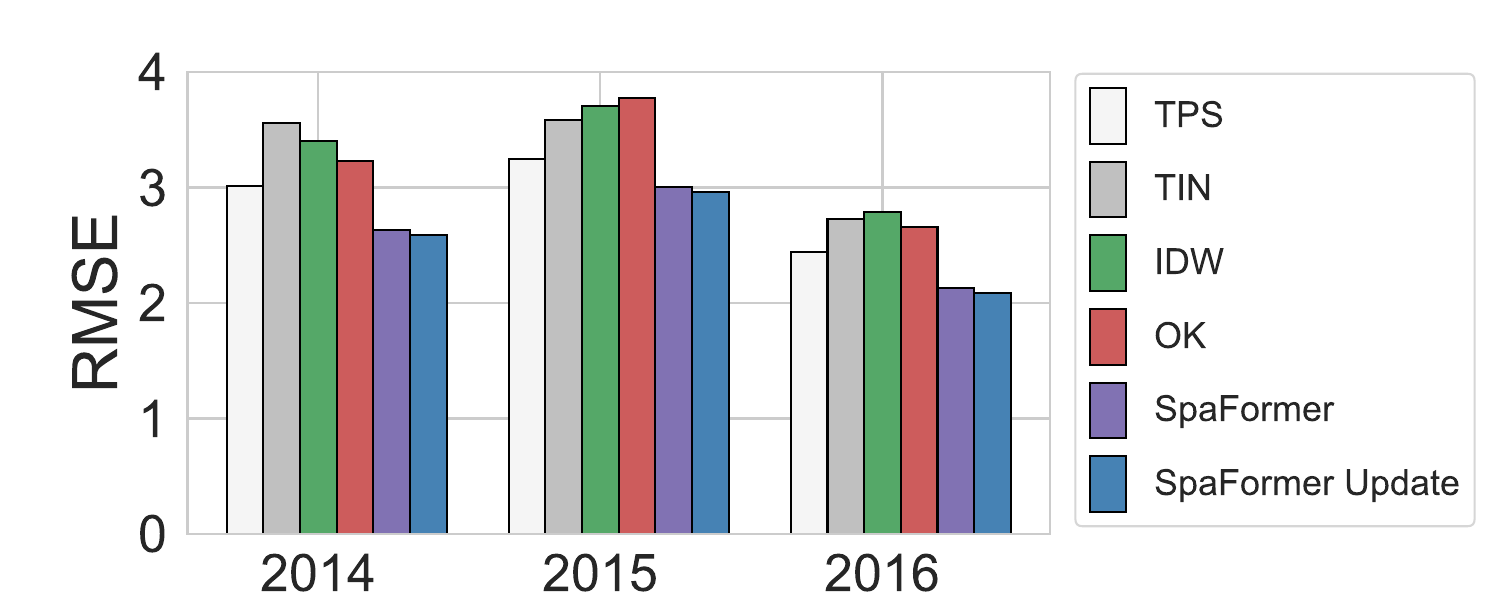}
\caption{The effect of model update on the HK dataset.}
\label{model_update}
\end{figure}

Over time, more rainfall data becomes available. However, different from the streaming data, rainfall does not happen all the time. Considering the intermittency of rainfall events, we can collect rainfall data and retrain the model at a low frequency (e.g., year by year) based on all the data.
Here, we take the HK region as an example, add new coming rainfall data into the training data and update the model year by year. The evaluation results for the 2014, 2015, and 2016 years are shown in Figure~\ref{model_update}, four traditional methods are added as a comparison.
Specifically, ``SpaFormer'' denotes the trained model on 2008-2012 training data, while ``SpaFormer Update'' is the trained model by adding new data\footnote{E.g., using the trained model on 2008-2013 training data to evaluate 2014 test data, using the trained model on 2008-2014 training data to evaluate 2015 test data.}.
From the results, we can see that our proposed method consistently outperforms other baselines. Besides, as more new data are added, the updated SpaFormer model can achieve lower errors than the old one.

\begin{table}[t]
\caption{The effect of training data amount.}
\resizebox{0.8\columnwidth}{!}{
\begin{tabular}{c|rrr|rrr}
\hline
\multirow{2}{*}{\textbf{Training Data Amount}} & \multicolumn{3}{c|}{\textbf{HK Dataset}}             & \multicolumn{3}{c}{\textbf{BW Dataset}}              \\ \cline{2-7} 
                                   & RMSE$\downarrow$ & MAE$\downarrow$ & NSE$\uparrow$   & RMSE$\downarrow$ & MAE$\downarrow$ & NSE$\uparrow$   \\ \hline
original        & 2.3328                               & 0.8329                              & 0.8520                             & 0.9874                               & 0.3278                              & 0.5158                            \\
$\times$ 2           & 2.2932                               & 0.8049                              & 0.8570                             & 0.9816                               & 0.3183                              & 0.5215                            \\
$\times$ 3           & 2.2846                               & 0.8024                              & 0.8581                             & 0.9797                               & 0.3139                              & 0.5234                            \\ \hline
\end{tabular}
}
\label{tab:data_amount}
\end{table}

\subsubsection{Transferability Study} \label{sec:Transferability}
In this section, we demonstrate the transferability of SpaFormer.
We apply the SpaFormer model trained on one dataset to the other dataset: the trained model on the HK dataset is transferred to BW, and the trained model on the BW dataset is transferred to HK.
Without fine-tuning, the trained model on the source dataset is directly applied to the test data of the target dataset. The results are shown in Table~\ref{tab:trans_direct}.
By comparing Table~\ref{tab:trans_direct} with Table~\ref{tab:main_result}, we can see that the SpaFormer model has good transferability between HK and BW datasets --- the model trained on each other outperform all other baselines except the SpaFormer trained on itself.
The transferability of SpaFormer confirms that such self-supervised learning based on spatial context is a promising solution to solving the spatial interpolation task: the transferred model can offer competitive results even when the new dataset is never seen. 

\begin{table}[t]
\caption{Transferability of SpaFormer model.}
\resizebox{0.8\columnwidth}{!}{
\begin{tabular}{c|ccc|ccc}
\hline
\multirow{2}{*}{\textbf{Methods}} & \multicolumn{3}{c|}{\textbf{HK Dataset}}             & \multicolumn{3}{c}{\textbf{BW Dataset}}              \\ \cline{2-7} 
                                   & RMSE$\downarrow$ & MAE$\downarrow$ & NSE$\uparrow$   & RMSE$\downarrow$ & MAE$\downarrow$ & NSE$\uparrow$   \\ \hline
SpaFormer          & \multicolumn{1}{r}{2.3328} & 0.8329                              & 0.8520                             & \multicolumn{1}{r}{0.9874} & 0.3278                              & 0.5158                            \\
SpaFormer Transfer & \multicolumn{1}{r}{2.4137} & 0.8581                              & 0.8416                             & \multicolumn{1}{r}{1.0007} & 0.3399                              & 0.5028                            \\ \hline
\end{tabular}
}
\label{tab:trans_direct}
\end{table}

\begin{table}[t]
\caption{Results on the PEMS-BAY dataset.}
\resizebox{0.6\columnwidth}{!}{
\setlength{\tabcolsep}{2mm}{
\begin{tabular}{c|ccc}
\hline
Methods   & RMSE$\downarrow$ & MAE$\downarrow$ & NSE$\uparrow$   \\ \hline
TIN       & 20.4678          & 10.1869         & -3.4126         \\
IDW       & 6.7235           & 3.7625          & 0.5239          \\
TPS       & 14.0928          & 7.2843          & -1.0919         \\
OK        & 8.2541           & 4.7571          & 0.2824          \\
KCN       & 8.0872           & 4.7568          & 0.3111          \\
IGNNK     & {\ul 6.1615}     & {\ul 3.6767}    & {\ul 0.6002}    \\ \hline
SpaFormer & \textbf{5.8954}  & \textbf{3.4818} & \textbf{0.6339} \\
Improv.   & 4.32\%           & 5.30\%          & 5.61\%          \\ \hline
\end{tabular}
}}
\label{tab:traffic}
\end{table}

\subsection{Other Case - Traffic Spatial Interpolation}
Although our solution is proposed for rainfall spatial interpolation, it can be applied to spatial interpolation problems in other domains thanks to its generality.
In this section, we take traffic spatial interpolation as a case to further explore the performance of our proposed method. Performing spatial interpolation on traffic data is helpful in inferring traffic conditions at road locations without sensors installed.
We employ one commonly used real-world dataset, PEMS-BAY, to evaluate the performance of different methods. PEMS-BAY is a traffic speed dataset in the Bay Area, released by~\cite{ml33}. Specifically, PEMS-BAY is recorded every 5 minutes and contains data from 325 sensors and 52,116 timestamps.
Same as the rainfall interpolation, we randomly sample 20\% sensors as the test set, and the rest are used as the training data.

It is worth noting that the correlation between traffic sensors is usually related to travel distance instead of geographic distance.
So here we make a simple modification in SpaFormer, that is, using the travel distance instead of the geographic distance to generate the relative position embedding.
Similarly, the inverse distance matrix of IDW, and the adjacency matrix of KCN and IGNNK are also calculated based on travel distances.
Due to the methodological limitation, the methods TIN, TPS, and OK cannot make use of the travel distance information, instead, they employ point coordinates as the inputs to fit the local surface by using deterministic polynomial functions or statistical relationships among points.

As shown in Table~\ref{tab:traffic}, SpaFormer consistently achieves the best performance in all metrics, which indicates the effectiveness and generality of our method.
Traditional methods TIN, TPS, and OK perform worst because they cannot make use of the travel distance information, and the better results of IDW also verify the importance of travel distance in traffic data.
KCN, a GNN-based method, suffers from the limitation of the model architecture and thus performs poorly. KCN builds a local subgraph around a central point and only predicts the value of the central point, which leads to a weak supervision signal and cannot capture global information on a sparse traffic graph due to a general two-layer design.
Instead, IGNNK and our proposed SpaFormer randomly mask multiple nodes and reconstruct their values, thus producing useful supervision signals to guide the node representation learning.
Compared with IGNNK, our method can adaptively capture the spatial correlations without pre-settings, thus achieving better results.

%% file: 5-Conclusions.tex
\section{Conclusions} \label{sec:Conclusions}
In this paper, we study the rainfall spatial interpolation task.
We present SSIN, a self-supervised learning framework, to train an effective spatial interpolator from the rich spatial patterns of historical data.
Specifically, we propose the SpaFormer model as the core of SSIN, which can accurately infer rainfall values of unobserved locations by learning informative embeddings and adaptively modeling spatial correlations based on spatial context.
The empirical study shows that our proposed SpaFormer consistently outperforms state-of-the-art methods in two large real-life raingauge datasets.
Besides, SpaFormer demonstrates remarkable transferability in the two raingauge datasets on different regions, which makes it possible to solve other spatial interpolation scenarios without sufficient training data.
Furthermore, we take traffic spatial interpolation as another use case and conduct additional experiments. 
The experimental results show that our solution significantly outperforms other baselines, which verifies its effectiveness and generality.